# Designing Interpretable ML System to Enhance Trust in Healthcare: A Systematic Review to Proposed Responsible Clinician-AI-Collaboration Framework

Elham Nasarian [1, a], Roohallah Alizadehsani[b], U. Rajendra Acharya[c,d], Kwok-Leung Tsui[a].

[a] *Grado Department of Industrial & Systems Engineering, Virginia Tech, Blacksburg, VA 24061, USA*

[b] *Institute for Intelligent Systems Research and Innovation (IISRI), Deakin University, VIC 3216, Australia*

[c] *School of Mathematics, Physics, and Computing, University of Southern Queensland, Springfield, Australia*

[d] *Center for Health Research, University of Southern Queensland, Springfield, Australia*

**Abstract:**

**Background:** Artificial intelligence (AI)-based medical devices and digital health technologies, including medical sensors, wearable health trackers, telemedicine, mobile health (mHealth), large language models (LLMs), and digital care twins (DCTs), significantly influence the process of clinical decision support systems (CDSS) in healthcare and medical applications. However, given the complexity of medical decisions, it is crucial that results generated by AI tools not only be correct but also carefully evaluated, understandable, and explainable to end-users, especially clinicians. The lack of interpretability in communicating AI clinical decisions can lead to mistrust among decision-makers and a reluctance to use these technologies.

**Objective:** This paper systematically reviews the processes and challenges associated with interpretable machine learning (IML) and explainable artificial intelligence (XAI) within the healthcare and medical domains. Its main goals are to examine the processes of IML and XAI, their related methods, applications, and the implementation challenges they pose in digital health interventions (DHIs), particularly from a quality control perspective, to help understand and improve communication between AI systems and clinicians. The IML process is categorized into pre-processing interpretability, interpretable modeling, and post-processing interpretability. This paper aims to foster a comprehensive understanding of the significance of a robust interpretability approach in clinical decision support systems (CDSS) by reviewing related experimental results. The goal is to provide future researchers with insights for creating clinician-AI tools that are more communicable in healthcare decision support systems and offer a deeper understanding of their challenges.

**Methods:** Our research questions, eligibility criteria, and primary goals were proved using the Preferred Reporting Items for Systematic Reviews and Meta-Analyses (PRISMA) guideline and the PICO (population, intervention, control, and outcomes) method. We systematically searched PubMed, Scopus, and Web of Science databases using sensitive and specific search strings. Subsequently, duplicate papers were removed using EndNote and Covidence. A two-phase selection process was then carried out on Covidence, starting with screening by title and abstract, followed by a full-text appraisal. The Meta Quality Appraisal Tool (MetaQAT) was used to assess the quality and risk of bias. Finally, a standardized data extraction tool was employed for reliable data mining.

[1] Corresponding Authors. Elham Nasarian. Grado Department of Industrial & Systems Engineering, Virginia Tech, Blacksburg, VA 24061, USA. Email: elhamn20@vt.edu

**Results:** The searches yielded 2,241 records, from which 555 duplicate papers were removed. During the title and abstract screening step, 958 papers were excluded, and the full-text review step excluded 482 studies. Subsequently, in quality and risk of bias assessment, 172 papers were removed. 74 publications were selected for data extraction, which formed 10 insightful reviews and 64 related experimental studies.

**Conclusion:** The paper provides general definitions of explainable artificial intelligence (XAI) in the medical domain and introduces a framework for interpretability in clinical decision support systems structured across three levels. It explores XAI-related health applications within each tier of this framework, underpinned by a review of related experimental findings. Furthermore, the paper engages in a detailed discussion of quality assessment tools for evaluating XAI in intelligent health systems. It also presents a step-by-step roadmap for implementing XAI in clinical settings. To direct future research toward bridging current gaps, the paper examines the importance of XAI models from various angles and acknowledges their limitations.

## 1. Introduction

Nowadays, artificial intelligence (AI) is deeply integrated into our lives, aiding various sectors in addressing complex challenges and revolutionizing traditional decision-making methods [1]. Machine learning (ML) and AI models can be found in many aspects of our daily lives, such as in cars for navigation and in smartwatches to track health [2], in smartphones to recognize our voices, in clinics to predict diseases [3], in hospitals to help physicians for surgery [4], in medicine for drug discovery [5], etc. Furthermore, various industries are exploring the implementation of AI solutions into their processes, given their substantial efficiency compared to traditional methods [6]. In healthcare and medicine, particularly in clinical decision support systems, the application of AI techniques is consistently evolving and progressing. Accurate predictions in an intelligent health system require vast amounts of data, and most electronic medical records (EMR) are multidimensional and extraordinarily complex [7]. Moreover, since classical machine learning algorithms, such as random forest (RF), decision tree (DT), and linear regression (LR), are less precise compared to deep neural networks (DNN) in extracting hidden patterns from clinical data, a complex DNN must be applied [8]. However, as the number of layers, features, and hyperparameters in these predictive models increases, the operation of these AI tools becomes progressively more difficult to understand. [9].

The training process in DNN methods is highly sophisticated due to regularization, hyperparameter tuning, and loss functions. Therefore, an output from DNN algorithms is more complex to understand and trust compared to outputs from classical ML methods like RF, DT, and KNN [10]. According to these results, there is a significant challenge for decision-makers and caregivers: the Black-Box model [11]. Basic ML models are clearer to interpret. In this context, interpretation refers to the ability to provide an easy-to-understand explanation aimed at enhancing clinician-AI communication within the clinical decision support system (CDSS) [12]. Simpler ML models are considered White-Box models, which do not require added parameters or functions to yield transparent results. Additionally, there exists a concept that falls between the Black-Box and White-Box models, known as the Gray-Box models. These methods can be easily explained if they are well-designed [13]. The following paragraphs will describe the challenges faced by black-box models in healthcare systems.

### *1.1.The general concept of interpretability in healthcare*

AI is an innovative technology that offers many benefits for decision-makers in real-world applications. However, modern AI systems face challenges in providing easily understandable explanations for their decisions due to the complexity of their algorithms. This opacity can lead to mistrust among end-users, especially in critical fields such as healthcare and medicine [14]. To address the Black-Box issue, developers must prioritize interpretability over accuracy and performance. This priority has led to the rise of Interpretable Machine Learning (IML) and Explainable Artificial Intelligence (XAI) in recent years. Interpreting a machine learning model involves understanding its predictions and the decision-making process for patients, clinicians, and developers. Interpretability is crucial in many contexts, whether it be for meeting legal requirements, preventing biased decisions, or boosting user confidence [15]. Interpretability offers several advantages, including (1) helping users to find clear patterns in ML models; (2) enabling users to understand the reasons behind inaccurate predictions; (3) building trust among end-users in model predictions; (4) empowering users to detect bias in ML models; and (5) providing an added safety measure against overfitting. In most cases, ML and DL methods function as Black-Box models, where users are unaware of the internal workings, how features are selected, and how predictions are made. This lack of transparency often leads to doubts and raises fundamental questions about the reliability of the models for decision-makers in health systems. Figure 1 illustrates frequent questions and concerns about the AI decision-making process.

XAI has many applications across healthcare and medicine. Currently, interpretability is the primary focus in the medical domain to provide easy-to-understand results for patients and caregivers and to enhance their trust in specific applications, such as drug discovery [16], thyroid [17], parental stress [18], wearable health trackers and biosensors [19], and respiratory disease [20]. The Black-Box issue in AI arises when a system faces challenges in providing a clear explanation for how the model arrived at a decision. The terms Black-Box, Gray-Box, and White-Box describe distinct levels of transparency about the inner workings of machine learning algorithms. The fundamentals of explainability are closely related to interpretability; methods are considered explainable if humans can understand how they work and make decisions.

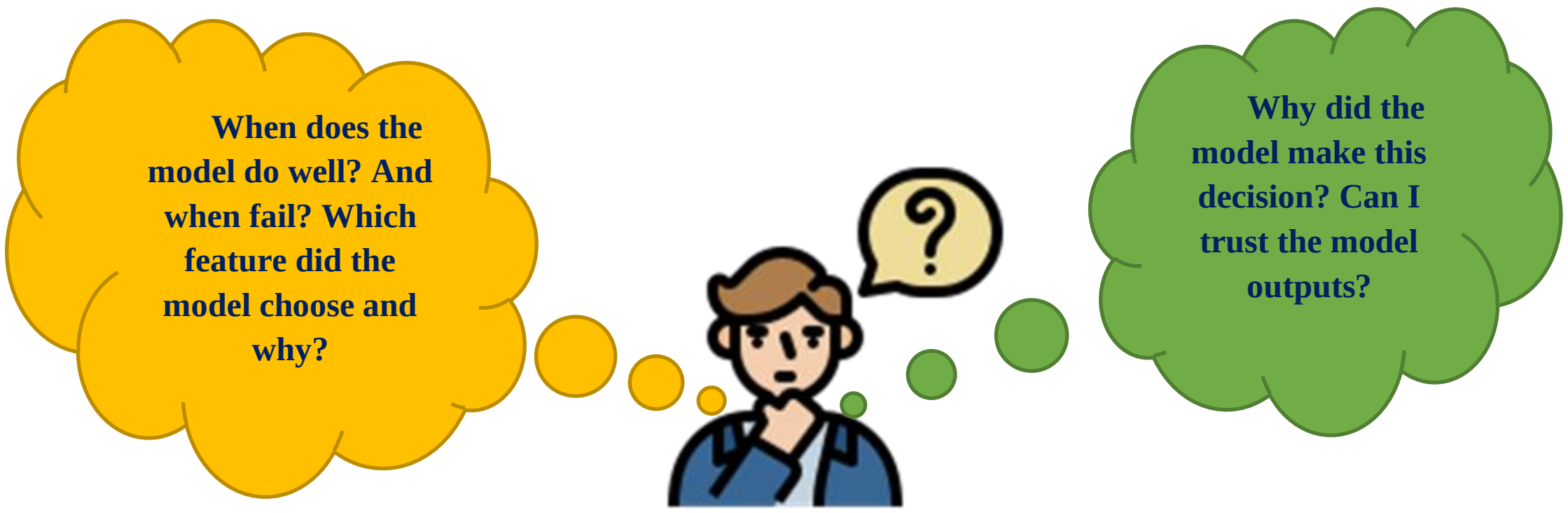

*Figure 1. Illustration of common concerns about AI in Healthcare, such as the circumstances of the model's success or failure, the reasons for feature selection, and the overall trustworthiness of the model's outputs.*

### *1.2. The Importance of transparency in healthcare*

Transparency and explainability are essential for AI implementation in healthcare settings, as inaccurate decision-making, such as disease predictions, could lead to serious challenges [11]. Caregivers and decision-makers in health systems are overly concerned about the Black-Box nature of AI tools. The approach involves creating ML models with a balanced trade-off between interpretability and accuracy, which is achieved by either (a) designing white-box or gray-box models that are inherently interpretable [21] while maintaining high accuracy, or (b) enhancing Black-Box models with a basic level of interpretability when White-Box or Gray-Box models cannot achieve an acceptable level of accuracy. While some researchers believe explaining Black-Box models can be beneficial, it is preferable to create interpretable models from the beginning of the development process [21]. Relying on explaining Black-Box tools, rather than inherently designing interpretable models, might lead to severe outcomes for patients in clinical settings. The interpretability method goes beyond simple predictions; it provides the extra information to understand how the AI clinical decision support system works. This is especially useful for end-users such as developers and clinicians [10]. Explainability, on the other hand, offers the end-user an understanding of the clinical decision-making process, which helps set up trust that AI tools are making exact and unbiased decisions based on information [1].

In other words, White-Box models are designed for interpretability, making their outcomes easier to understand but may result in slightly lower accuracy. They are preferred for applications where interpretability and transparency are critical, such as in medical diagnoses [14]. On the other hand, Black-Box models, while more correct, are less interpretable. They are preferred for tasks where achieving the highest level of predictive accuracy is the primary goal and interpretability is less critical, such as in image recognition for criminal issues. Gray-Box models provide a balance, offering a good trade-off between interpretability and accuracy. They can be suitable for a wide range of applications, providing a middle ground between white and black-box models [22]. Figure 2 shows the difference between these models.

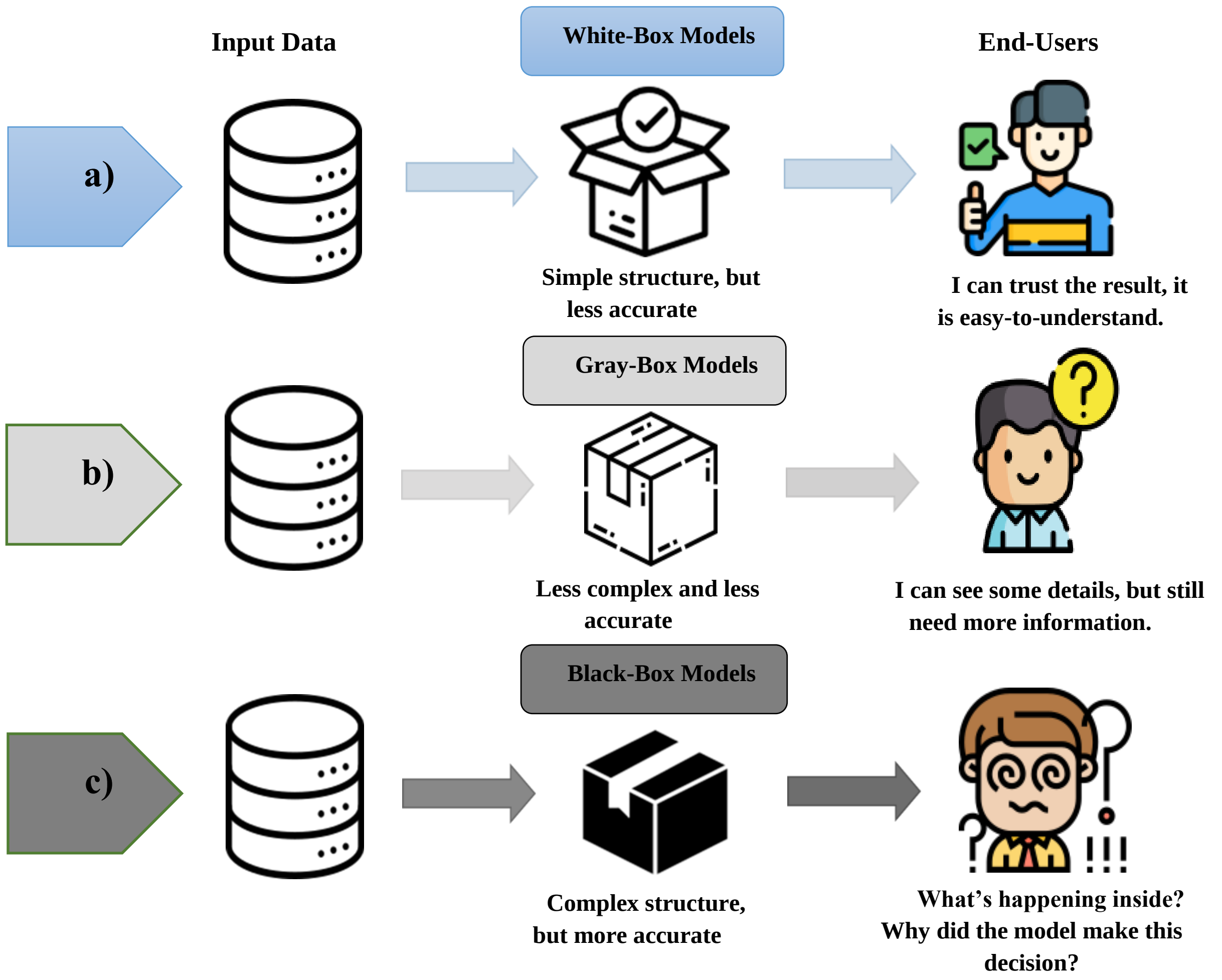

*Figure 2. A Comparison of a) White-Box, b) Gray-Box, and c) Black-Box models: illustrates the trade-offs between different types of machine learning models: a) White-Box models are fully transparent and understandable, garnering user trust despite potentially lower accuracy, b) Gray-Box models provide a balance with some explainability and moderate complexity, though end-users still seek further clarity, and c) Black-Box models offer high accuracy but lack transparency, leading to end-user confusion about decision-making processes.*

### *1.3. Balancing accuracy and interpretability in healthcare systems*

In healthcare and medicine, researchers strive to find the right models, while clinicians seek easy-to-understand and interpret models. However, prioritizing achieving the highest accuracy can result in more complex and harder-to-interpret models. Recognizing this balance is crucial for effective healthcare decision-making, especially in areas like predictive analytics for chronic diseases, as AI becomes increasingly prevalent in the medical domain. In some experiments, combining interpretable models can offer more explainable insights; however, some level of interpretability may be sacrificed to achieve maximum accuracy. For instance, while DTs (Decision Tree) are inherently interpretable, when they are used repeatedly and combined in a model like RF, they become difficult to understand. This underscores that sophisticated models like DNN may compromise interpretability for greater accuracy. In this context, explainability becomes more critical because it aids in understanding complex Black-Box systems [22]. For instance, deep learning techniques like CNN (Convolutional neural networks) are less interpretable than RF, and classical ML methods like DT are more explainable than RF. While classical ML models like LR have fewer functions, making them easier to understand, the model becomes highly complicated as the number of parameters increases. [23]. Defining a clear boundary between Black-Box, Gray-Box, and White-Box models is challenging.

It is significant to highlight that some authors note there is no scientific evidence supporting a general balance between accuracy and interpretability [23]. While most ML models, in their pursuit of achieving the highest accuracy, tend to sacrifice interpretability, designing for precise interpretability from the pre-processing stage can lead to a significant trade-off between interpretability and performance. Additionally, given the critical importance of healthcare decision support systems and their potentially life-changing impacts on individuals, it is essential to implement an interpretable system throughout the entire process of intelligent health systems, from data collection to decision-making. Explanations are vital for understanding the entire process, enhancing trust between end-users and the system, improving interaction and communication between clinicians and AI, aiding the decision support system in learning and updating over time, and easing comparisons between different ML models [23]. The importance of IML and XAI implementation in healthcare and medicine has been increasingly highlighted by researchers in recent years [24], [16]. This study offers a comprehensive overview of the significance of the interpretability process in clinical decision support systems, encompassing pre-processing interpretability, interpretable modeling, and post-processing interpretability. Explainability transcends academic interest; it will become a crucial aspect of future AI applications in healthcare and medicine, affecting the daily lives of millions of caregivers and patients.

### *1.4.Role of explainability in digital health interventions (DHIs)*

XAI tools aim to make intelligent health systems more communicable and transparent to digital health interventions (DHIs), including patients, their families, healthcare professionals, health system managers, and data services, following WHO (World Health Organization) guidelines [25]. Automated clinical decision-making and problem-solving systems may understand complex structures in medical multidimensional data but suffer from explaining hidden patterns in Black-Box models. This demonstrates that AI brings creative solutions and also presents critical challenges such as security, privacy, inclusion and diversity, and transparency simultaneously [26]. Figure 3 highlights benefits by opening a window to implementing XAI in the DHIs domain.

The classification of DHIs should be used in implementing intelligent health systems to address health needs, such as interpretable decisions, thereby enhancing clinician AI communication. This framework offers an overview of needs and challenges to aid decision-makers in emphasizing the significance of explainability and interpretability for eHealth outcomes [10]. This classification of DHIs also describes methods suitable for various health systems, such as electronic health records (EHR), telemedicine platforms, and clinical decision support systems (CDSS). According to WHO guidelines aimed at enhancing high communicability in the interpretability process of AI, it is crucial to consider all these DHIs (digital health interventions) categories in medical digital solutions as transmission and communication systems between targeted and untargeted clients. Additionally, they serve as decision support, consulting, and communication systems between healthcare providers [27].

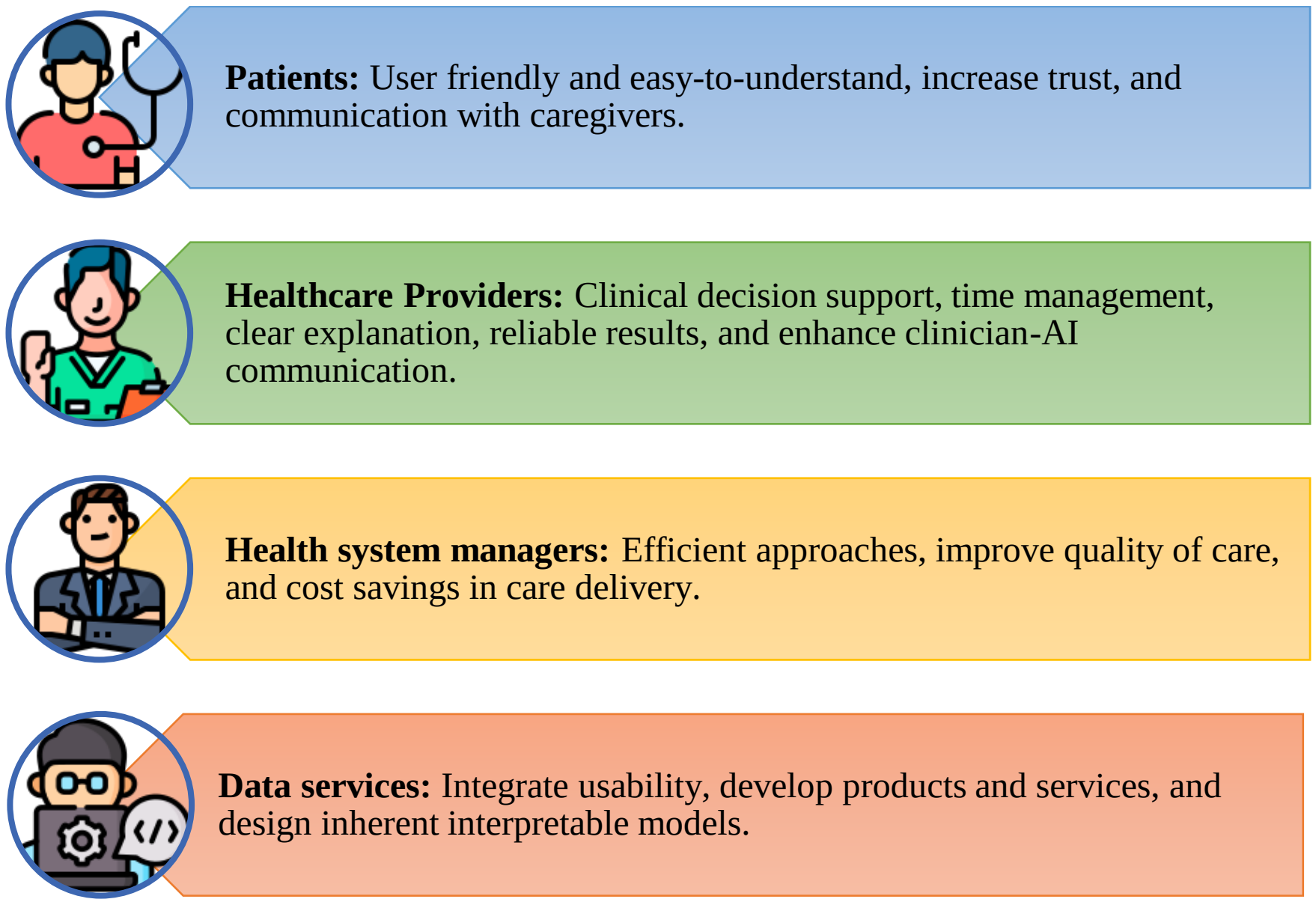

*Figure 3. Role of XAI in digital health interventions: presents the benefits of XAI across different stakeholders in digital health: improving patient trust, aiding healthcare providers in decision-making, enhancing system efficiency for managers, and fostering product development in data services.*

### *1.5. Current approaches of interpretability in health application*

There are two main explanations for ML decisions: inherent explainability and post-processing explainability approaches. Inherent explainability refers to the connections between straightforward and easily understandable input data and the model's outputs. Models like DT or LR exemplify this approach. However, even simple, easy-to-understand models can be limited by hidden factors that are not at once recognized. Studies in human-computer interaction have revealed that excessive transparency can hinder users' ability to name and rectify significant model errors. This may occur because users become overwhelmed with information. This phenomenon holds true even for intentionally designed models to be clear and easily explainable. Furthermore, more research has proved that even data scientists can sometimes overly rely on interpretability tools, leading to difficulties in accurately describing visualizations generated by these tools. Thus, while interpretability is crucial for understanding and trusting machine learning models, achieving the right balance of transparency without overwhelming users is still a significant challenge [28].

Unlike inherently explainable models, many modern AI applications involve data and models that are too complex and multidimensional to be easily understood through a simple relationship between inputs and outputs. Examples of such complex models include those designed for tasks like image analysis, text processing, and sound recognition. In these cases, the focus has shifted towards elucidating the model's decision-making process, a practice known as explainability. While popular in medical imaging as a post-hoc explainable tool, saliency maps have limitations.

They may highlight areas holding a mixture of useful and less useful information. Moreover, they do not offer precise insight into what aspects the model considers important for diagnosis. For example, it is still unclear whether the model's decision is guided by a specific anomaly, shape, or technical detail [29].

Explainability methods in machine learning often face an interpretability gap because they rely on humans to figure out the meaning of a given explanation. However, humans tend to give a positive interpretation, assuming the feature they find important is the one the model used. This tendency is a well-known cognitive error called confirmation bias. Computer scientist Cynthia Rudin succinctly captures this issue by advocating for a shift away from explaining Black-Box ML models for high-stakes decisions and promoting the use of interpretable models. This approach aims to reduce the reliance on human interpretation and mitigate the potential biases introduced by confirmation bias. By employing inherently interpretable models, such as DT or LR, stakeholders can have more confidence in understanding and trusting the decision-making process without the need for complex explanations [21].

In addition to heat maps, various other methods have been developed to explain complex medical data. These include techniques like feature visualization. Feature visualization involves creating synthetic inputs that strongly activate specific parts of a machine learning model. This enables each model decision to be understood as a combination of features detected in the input. However, a limitation of this approach is that synthetic inputs often do not directly correspond to easily interpretable human features. Therefore, they face similar interpretability concerns as heat maps. For instance, if a synthetic input resembles a feature that a human would use to plan (such as a fur-like texture in a dog-detecting AI model), a human still needs to figure out if this implies the model made a good decision. In other words, while feature visualization provides insights into how the model works, the interpretation of these synthetic inputs is still subjective and requires careful consideration to ensure meaningful insights [30]. Similar concerns apply to other well-known current explanation methods like locally interpretable model-agnostic explanations (LIME) and Shapley values (SHAP). LIME focuses on understanding individual-level decisions by slightly changing the input example and finding which alterations are most likely to affect the decision. Image analysis involves occluding parts of the image, resulting in a heat map that describes the crucial components of the decision. However, these explanations face similar interpretability challenges as saliency mapping. Both LIME and SHAP are versatile and used across diverse types of healthcare data, including structured data from electronic health records and electroencephalogram waveform data [31] [32].

### *1.6. Motivations and aims of this review*

In recent years, scholars in healthcare and medicine have investigated the role of XAI and IML in real-world medical applications, such as disease diagnosis. However, most existing reviews and surveys in this field primarily focus on comparing the differences between various XAI methods [24], exploring the benefits of White-Box models [33], talking about reasons for avoiding Black-Box models [34], as well as looking at related guidelines in health systems [26]. Some reviews highlighted the role of XAI in specific health applications such as oncology [35], heart disease [36], and drug discovery [16]. Appendix 4 summarizes important surveys on and XAI in healthcare and medicine. In healthcare and medicine, it is crucial to interpret and understand AI models throughout the entire process of clinical decision support systems, including the pre-processing, processing, and post-processing stages. However, based on our current knowledge, there has been no review that has thoroughly examined the implementation of XAI and IML from the pre-processing stage. Most existing studies have primarily focused on explaining models and their outcomes. Additionally, it is essential to use relevant explanation tools during the pre-processing phase before delving into the selection of White-Box or Gray-Box models or trying to understand Black-Box models in predictive analysis and decision-making for patients. These tools can provide insights into the distribution, quality, and relationships within the data, aiding in identifying potential issues or biases that may affect model performance. By understanding the data landscape upfront, healthcare professionals can make more informed decisions about model selection and interpretation, improving the reliability and effectiveness of AI-driven healthcare solutions.

In the context of explainability in AI, researchers face a significant challenge: explanations often lack performance guarantees. Their performance is seldom assessed rigorously, and when tested, it often relies on heuristic measures rather than directly considering the human perspective. This poses a problem because explanations only approximate the model's decision process and may not fully represent how the underlying model behaves. Using post-processing explainable AI tools to evaluate model decisions introduces an added potential source of error. The explanation generated can be correct or incorrect, just like the model. Given these challenges, researchers must consider whether to favor the full, complex model, which may be beyond human understanding but has high, validated performance, or opt for an explanation mechanism that could lead to reduced and unvalidated accuracy. Explainability methods, while not useless, do have limitations. They can sometimes confuse or mislead when understanding complex AI behavior. When they describe how a model behaves, they cannot always justify its decisions. This gap causes relying on intuition, which can introduce biases. Over-reliance on explanations can lead to less careful oversight of AI systems. Detecting and preventing algorithm biases is crucial, but current research efforts fall short. Instead of using explanations for every prediction, it is better to view them as offering all views of how a model works. For example, if heat maps consistently show a diagnostic model focusing on irrelevant areas, it may signal a problem with the test data. As an illustration, in the case of a skin cancer detection model, heat maps revealed that it was paying more attention to surgical markings than the actual skin lesions, highlighting potential issues with the model's training or data quality [37].

Explanations have also helped discover new insights in fields like ophthalmology and radiography. The collective behavior of explanations offers valuable insights, rather than depending solely on one for a specific prediction [38]. Explanations alone may not give all the answers, but that does not imply blind trust in AI predictions. It is imperative to meticulously confirm AI systems for safety and effectiveness, akin to the evaluation process for drugs and medical devices. Employing rigorous evaluations, particularly through randomized controlled trials, is crucial. Moreover, in cases like investigating racial bias in AI, transparency alone may not suffice. A comprehensive analysis of inputs, outputs, and outcomes is necessary to uncover any biases. Explanations can serve as a valuable tool for analysis, especially for developers, auditors, and regulators of AI systems. They are not solely for end users or subjects of AI. This paper presents a systematic review of the processes of IML and XAI and their application in clinical domains. Additionally, it provides details on XAI applications in digital health technologies such as medical sensors, wearable health trackers, and large clinical language models. We also propose a step-by-step implementation roadmap and discuss key challenges and future directions for implementing IML in intelligent health systems. The primary goals of this paper are to review the process of IML and XAI, related methods, applications, and their implementation challenges in the context of healthcare and medicine, particularly in CDSS. The interpretability process is classified into pre-processing interpretability, interpretable modeling, and post-processing interpretability. Significantly, the paper explores the role of IML in healthcare and medical issues, emphasizing its necessity in digital health solutions. The paper aims to comprehensively understand the importance of interpretability in health decision-making systems by reviewing related experimental results. The goal is to provide future researchers with insights for the more reliable implementation of AI tools in healthcare and a deeper understanding of the challenges they might face. To our knowledge, this paper presents the first systematic review of the entire process of interpretability and explainability in intelligent health systems, accompanied by a step-by-step evaluation and implementation roadmap. For this reason, we have named 5 research questions using the PICO framework [39], [40]:

1. Which levels of interpretability have been applied in health applications?
2. Which healthcare applications have used interpretable AI tools?
3. What are the current approaches to interpretability in health applications?
4. How can we implement and evaluate robust interpretable AI for specific end-users?
5. What are the potential future trends and key challenges in the medical domain using explainable AI?

The structure of the paper is shown in Figure 4.

## 2. Systematic review method

The recommended reporting elements for systematic reviews and meta-analyses (PRISMA) criteria were followed in conducting the current systemic review [39, 40]. The search technique, selection standards, and data extraction were all included in the review protocol. The following subsection explains this part:

### *2.1.Search strategy and selection criteria*

As the field of digital health and remote care monitoring in the clinical decision support systems setting continues to evolve, researchers have aimed to track its advancements by conducting various surveys and experimental results, our effort is to stay updated on this progress.

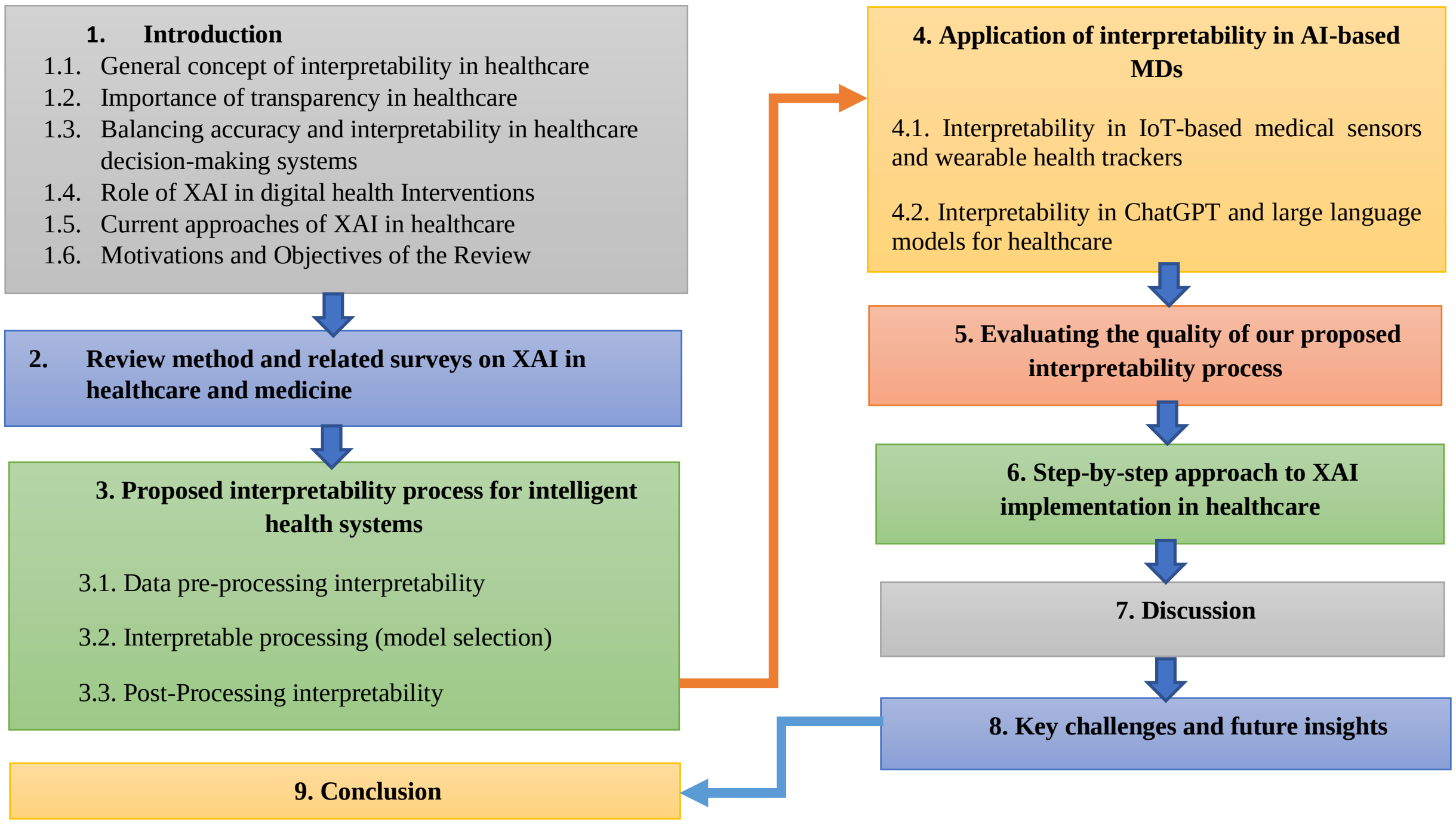

*Figure 4. Structure of the paper.*

In this systematic review, we tried to search for IML and XAI tools and follow their three steps process that propose solutions for improving transparency, trust, explainability, and interpretability for IoT-based sensors, wearables, telemedicine, mHealth, medical LLMs, and digital care twins. For this reason, it was started by applying rapid synthesis of evidence with data mining via a few keywords in PubMed, Scopus, and Web of Science databases such as "Explainable AI," "Interpretable ML," "Wearables," and "Telehealth." This exercise served as a preliminary measure of exploring research activities in the field of interest, guiding a more targeted search by showing arbitrary review boundaries within the study time and practical constraints. Therefore, we derived nine conceptual frameworks to guide structured searching for studies on "Interpretable Remote Health Monitoring in intelligent health system": Internet of medical things, telehealth, clinical decision support system, remote care, sensors, wearable medical devices, medical LLMs, digital care twins, explainable AI, and interpretable ML. Conducted in PubMed, Scopus, and Web of Science, we designed search strings for each conceptual framework based on title/abstract and specific term. The final Boolean search strings for this study were conducted on 16 June 2023. Following the data mining of evidence, by using the PRISMA guideline and PICO method

As digital health and remote care monitoring in the clinical domain evolves, researchers have aimed to track its advancements by conducting surveys and experimental studies. In this systematic review, we tried to search for IML and XAI tools and follow their three-step process to propose solutions for improving transparency, trust, explainability, and interpretability in IoT-based sensors, wearables, telemedicine, mHealth, medical Large Language Models (LLMs), and digital care twins. To start this review, we began by applying rapid synthesis of evidence with data mining via a few keywords in PubMed, Scopus, and Web of Science databases, such as “Explainable AI,” “Interpretable ML,” “Wearables,” and “Telehealth.” This preliminary measure served as an exploration of research activities in the field, guiding a more targeted search within the time and practical constraints of this study. Subsequently, we derived nine conceptual frameworks to guide structured searching for studies on “Interpretable Remote Health Monitoring in intelligent health systems”: Internet of medical things, telehealth, clinical decision support system, remote care, sensors, wearable medical devices, medical LLMs, digital care twins, explainable AI, and interpretable ML. These frameworks aided in designing search strings for each conceptual framework based on title/abstract and specific terms. The final Boolean search strings for this study were conducted on June 16, 2023. Following the data mining of evidence, we used the PRISMA guideline and PICO method for analysis and synthesis [39], [40], We proved criteria to define the focus and boundaries of this systematic review. The selection eligibility criteria are summarized in

Appendix 2. Selections were conducted by independent reviewers. All titles and abstracts of included studies were imported into the reference manager Endnote21 to remove duplicates and then imported into Covidence. Reviewers initially excluded irrelevant studies based on title and abstract. If there was any doubt about exclusion, studies were accepted for full-text screening. Any discrepancies between reviewers were resolved through consensus.

### *2.2. Quality appraisal, risk of bias assessment and search for grey literature*

After full-text reviewing, the MetaQAT tool was employed for quality and risk of bias assessment. This tool functions as a meta-tool for evaluating public health evidence, aiding users in correctly applying various critical appraisal tools. It provides a larger framework for guidance and is specifically designed for public health evidence [41]. In this phase, all studies with a substantial risk of bias were excluded. This approach was adopted as a measure of good practice to ensure that only studies with robust methodological quality were included, thus avoiding potentially misleading readers about the reliability of the included studies. Figure 5 shows the process of this critical appraisal framework. Gray literature was considered not applicable to this study, which aimed to systematically review research activities to achieve a broad perspective on the evidence in this field. Following the screening of titles and abstracts, full-text reviewing, and assessment of the risk of bias and quality, all details of the final included studies and the results of the selected papers were extracted into a standardized data mining template in Microsoft Excel, keeping a structured format.

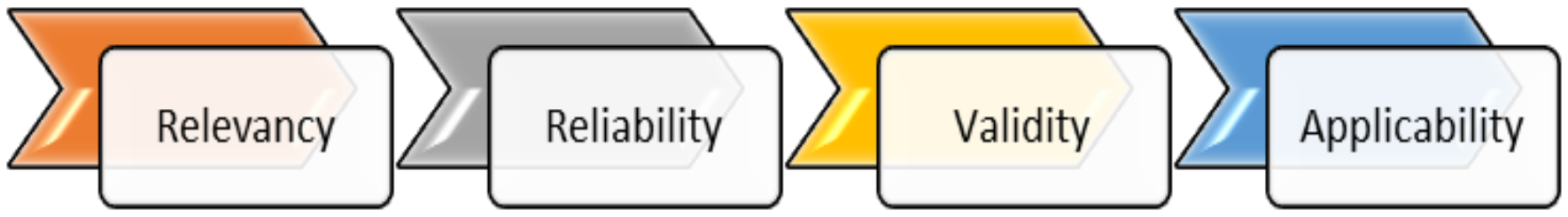

*Figure 5. Appraisal framework of MetaQAT tool: each symbolizes a critical assessment phase in the tool's method.*

### *2.3.Data analysis and search results*

The PRISMA 2020 flowchart for the selection process and systematic mapping of all included studies, based on standardized data extraction tools, was constructed in Covidence. Data were analyzed and narratively summarized using descriptive statistics in tables or graphs for each study aim [39], [40]. The structured searching via search strings based on conceptual frameworks yielded 658 studies in PubMed, 1001 articles in Scopus, and 582 publications in Web of Science. The final Boolean search operation revealed 2241 studies imported into Covidence. From these 2241 studies, 555 duplications were found. However, 958 studies were removed during the first selection stage via title and abstract. Subsequently, 482 studies were excluded following the second stage selection by full-text reviewing. Further, 172 papers were removed following the third step by critical appraisal and risk of bias assessment. 74 papers, including 10 related review papers and 64 experimental results publications, were selected. Figure 6 summarizes the PRISMA 2020 flow diagram for the final 74 studies for data extraction, including 10 existing reviews and 64 research papers.

### *2.4.Data mapping of included studies*

The standardized data extraction tool was employed to map data from all 74 studies included in this systematic review. This helped a comprehensive examination of all eligibility criteria outlined in the study. Appendix 4 and Appendix 5 present the results of this data mapping process; Appendix 4 delineates reviews, and Appendix 5 describes the experimental studies. The two authors meticulously examined the selected papers, with the investigation primarily focusing on the main approaches used during the process of interpretability in intelligent health systems. The explainability of the outcomes of the reviewed papers is verifiable, as most of these papers have been published in top-ranked journals and conferences. Furthermore, many reviewed papers on explainable AI and interpretable ML tools have already been used in practice, proving their capabilities.

### *2.5.Related surveys on explainable AI in healthcare and medicine*

Although the number of studies on XAI for real-world health applications is rapidly increasing (as shown in Appendix 4), there is still a gap between comprehensive surveys of the entire process of interpretability from data pre-processing to post-modeling explainability in clinical decision support systems, as well as an XAI evaluation and implementation framework for intelligent health systems. Several types of review papers on XAI in medical settings exist. Still, most of these surveys focus on comparing diverse types of XAI, its application, challenges, and related guidelines in healthcare and medicine. For instance, in [42], the authors first discuss the definition of AI-based medical devices and then compare existing XAI guidelines for stakeholders. Similarly, in [24], the authors emphasize the importance of XAI models in healthcare, highlight distinct types of XAI methods and their applications in the healthcare system, and provide insights into important challenges and future directions for XAI in clinical contexts.

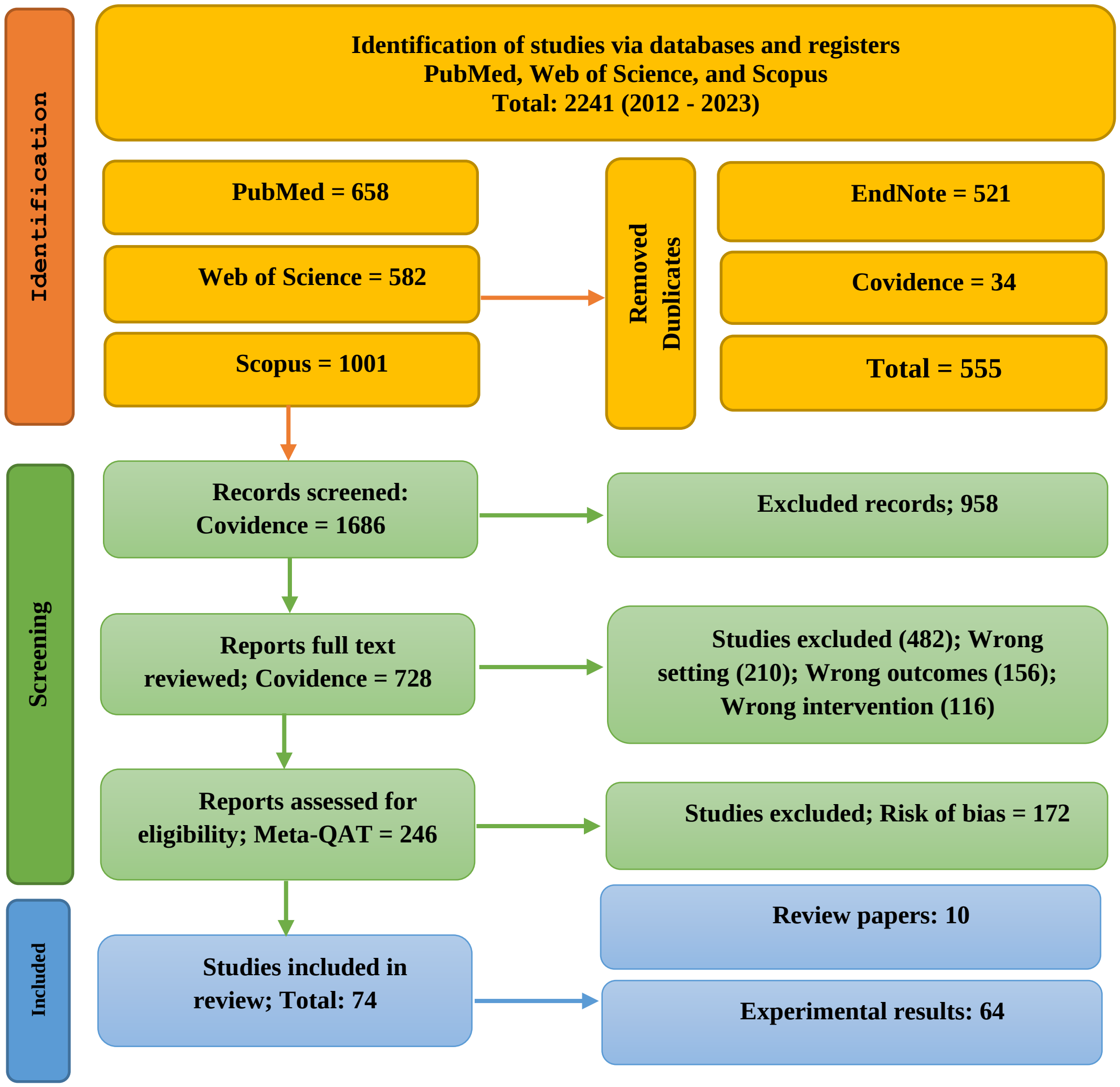

*Figure 6. PRISMA flow diagram of the selection process for a systematic review, showing the filtration of studies from identification through databases like PubMed, Web of Science, and Scopus, screening, and eligibility assessment, to the final inclusion of studies in the review.*

Moreover, some of these review studies have focused on specific applications. For instance, in [35] the authors compared diverse XAI methods, evaluating their benefits and current challenges to assess interpretability in precision oncology. Also, in [16], a comprehensive study was conducted on XAI methods, their application, and future insights for researchers and decision-makers in the field of drug discovery. For heart disease classification in [36], a comprehensive systematic review was conducted on interpretability methods, focusing on heart disease diagnosis from an Electrocardiogram (ECG) signal. [43] attempted to focus on Black-Box models, their limitations, and opportunities in cardiology applications to underscore the importance of XAI models in this field. More details and information about the most important existing surveys on XAI and IML in healthcare and medicine can be found in Appendix 4.

As Figure 7 illustrates, to the best of our knowledge, none of these reviews have conducted a comprehensive study on the entire Interpretable Machine Learning (IML) process, evaluation, and implementation framework, as well as several types of AI-based medical devices. Furthermore, most of these surveys have been published in the current year, proving the increasing importance of explainability and interpretability for proving a transparent, trustworthy, easy-to-understand, and reliable clinician-AI communication system for end-users. This figure highlights the comparison of our study with some important review papers published in the field of XAI for healthcare and medicine. Figure 7 provides a qualitative summary of the review papers included in our paper. This figure presents two visual summaries: a) A world map categorizes regions based on their contribution to review papers, with different shades showing the contribution level and b) A pie chart shows the percentage of review papers published from 2020 to 2023, with the largest share in 2023.

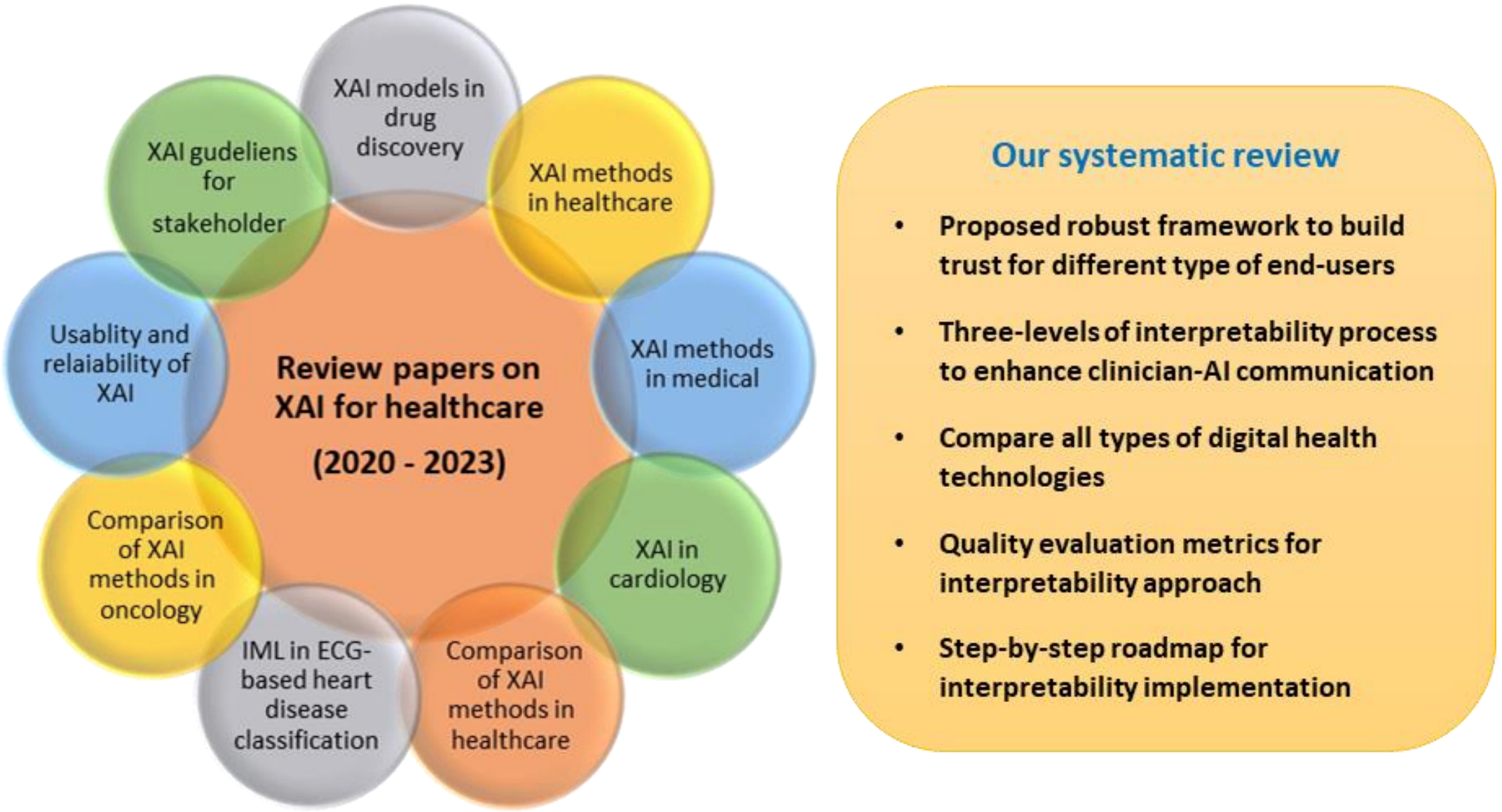

*Figure 7. Comparison of our study with other review papers of XAI in healthcare from 2020 to 2023, with a focus on the comprehensive coverage of the systematic review including a robust framework, interpretability processes, and a step-by-step roadmap for implementation.*

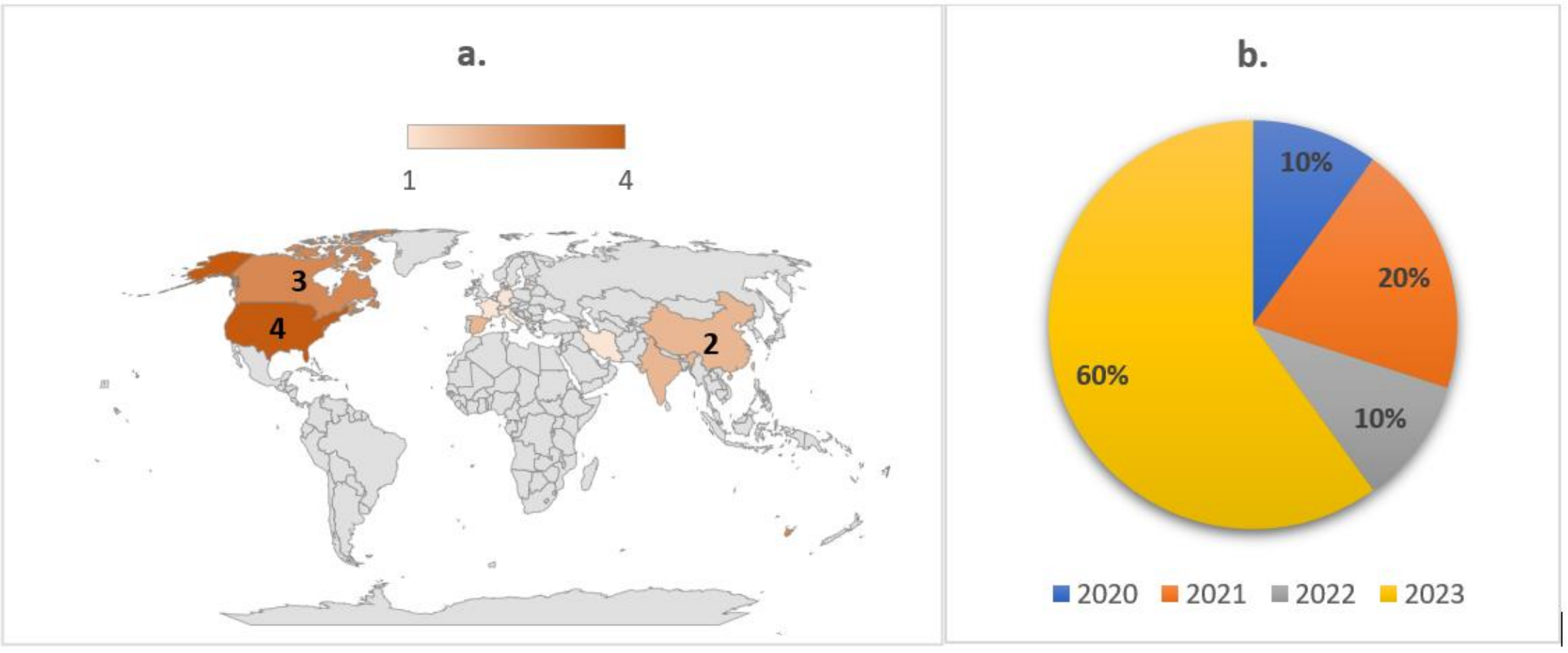

*Figure 8. Summary of reviews included: (a) Authors country-wise; a map displaying regions numbered 1 to 4 to indicate the volume of review papers associated with each. (b) year-wise distribution: a pie chart showing the yearly distribution of papers from 2020 to 2023, with the largest segment in 2023.*

## 3. Proposed interpretability process for intelligent health system

How interpretable should the explanation be to enhance communication between AI and clinicians? How interpretable should the explanation be in addressing end-user requirements? Since several factors can affect ML model decisions, a diverse range of user-friendly and trustworthy explanations throughout the entire process of clinical systems is needed. Explainability aims to understand the predictive models, from data collection and training processes to outcomes. While there is hope that merely explaining these AI models would be reasonable and understandable for end-users, this notion should be reconsidered. It is significant to design and implement inherently interpretable techniques from the preliminary stages in our process of decision support systems [21]. Nowadays, AI plays an essential role in healthcare and medicine domains. However, AI tools must be interpretable and transparent for end-users. End-users encompass anyone who uses or is affected by an ML model, including caregivers, clinicians, patients, users, data scientists, regulatory bodies, domain experts, executive board members, and managers [22]. Therefore, due to significant concerns about understanding AI tools' behavior, the implementation of AI in medical settings has been limited [14].

For healthcare decision-makers, AI systems must be reliable, exact, and transparent [33]. While many researchers have focused on using XAI after model selection and in the post-modeling stages, it is equally important to consider interpretability throughout the entire process: (1) data pre-processing, (2) during model selection, and (3) post-modeling. Focusing on only one of these aspects may not provide a clear understanding. Data scientists and developers may benefit more from understanding the internal workings of the model for performance improvement and preventing overfitting rather than post-processing interpretability. On the other hand, caregivers like physicians, nurses, and patients are more interested in understanding how and why a specific result was generated, along with the key features that influenced that decision. Therefore, it is critical to tailor explanations to all end-users and the entire decision-making process in the clinical setting.

For this purpose, we have introduced an interpretable process for healthcare applications, including pre-processing interpretability, interpretable processing, and post-processing interpretability. Figure 9 highlights our proposed interpretability process, and Appendix 5 summarizes the interpretable process in existing research papers. This study employs a novel framework that includes all three interpretability aspects. This approach offers two significant benefits: (1) It makes the design and development of explainable systems more cost-effective and easier to understand, and (2) It enhances satisfaction for end-users such as clients, patients, health providers, developers, researchers, and managers, as they receive a more focused and clear explanation tailored to their specific needs, rather than a generalized one. Researchers aim to classify explanation methods used to understand the reasoning behind learning algorithms, and they try to answer this question: what is the purpose of the interpretability methods? To provide a comprehensive explanation, interpretability techniques must address fundamental questions such as why and how the model generates predictions and makes decisions, as explored by many studies [44]. Error! Reference source not found. proves what we are looking for in each step of our proposed framework for the interpretability process in health systems. In the following section, we will discuss three levels of our proposed interpretability process and its applications in healthcare and medicine based on the research papers in this systematic review. Moreover, in Appendix 5, the included studies are summarized.

*Table 1. Summary of goals addressing 3 levels of the proposed IML process in healthcare.*

| Data Pre-Processing Interpretability | Interpretable Model Selection Process | Post-Processing Interpretability |
|---|---|---|
| **Pre1.** What type of medical data do we have? Is it high-dimensional? Unstructured? High-volume?<br>**Pre2.** Why did you collect these features for these patients? What are we looking for in this data?<br>**Pre3.** What are the key features of this data?<br>**Pre4.** Is clinical data clean and balanced?<br>**Pre5.** Can we reduce the number of features to achieve high accuracy?<br>**Pre6.** Do we need any data augmentation methods for better performance?<br>**Pre7.** Can we use feature engineering methods for better explanation in health apps?<br>**Pre8.** How can we use any visualization tools to better understand features related to patients? | **Mod1.** How can one consider a feature or parameter important in predictive analysis, like disease detection?<br>**Mod2.** What are the results of applying several methods and making different decisions?<br>**Mod3.** What is the association between these selected features and models?<br>**Mod4.** What are we looking for? Interpretability, performance, or both?<br>**Mod5.** Which metrics should we use to test and confirm the selected model?<br>**Mod6.** How can end-users make models more understandable?<br>**Mod7.** How can we use any visualization tools to better understand predictive models?<br>**Mod8.** How can we make this model more robust? | **Pos1.** What is the reason behind the decision made by the model? Did you consider ethics and bias?<br>**Pos2.** Which features had a high effect on the models' prediction?<br>**Pos3.** Does it have user-friendly outcomes for patients and clinician-AI communication?<br>**Pos4.** What if the information is changed? Did you consider any continuous improvement in this model?<br>**Pos5.** To keep current performance, what criteria must be met? What should we do to receive different results?<br>**Pos6.** What did we look for? Why is it important to come to a certain conclusion or decision?<br>**Pos7.** How can we correct errors? When does the model fail? |

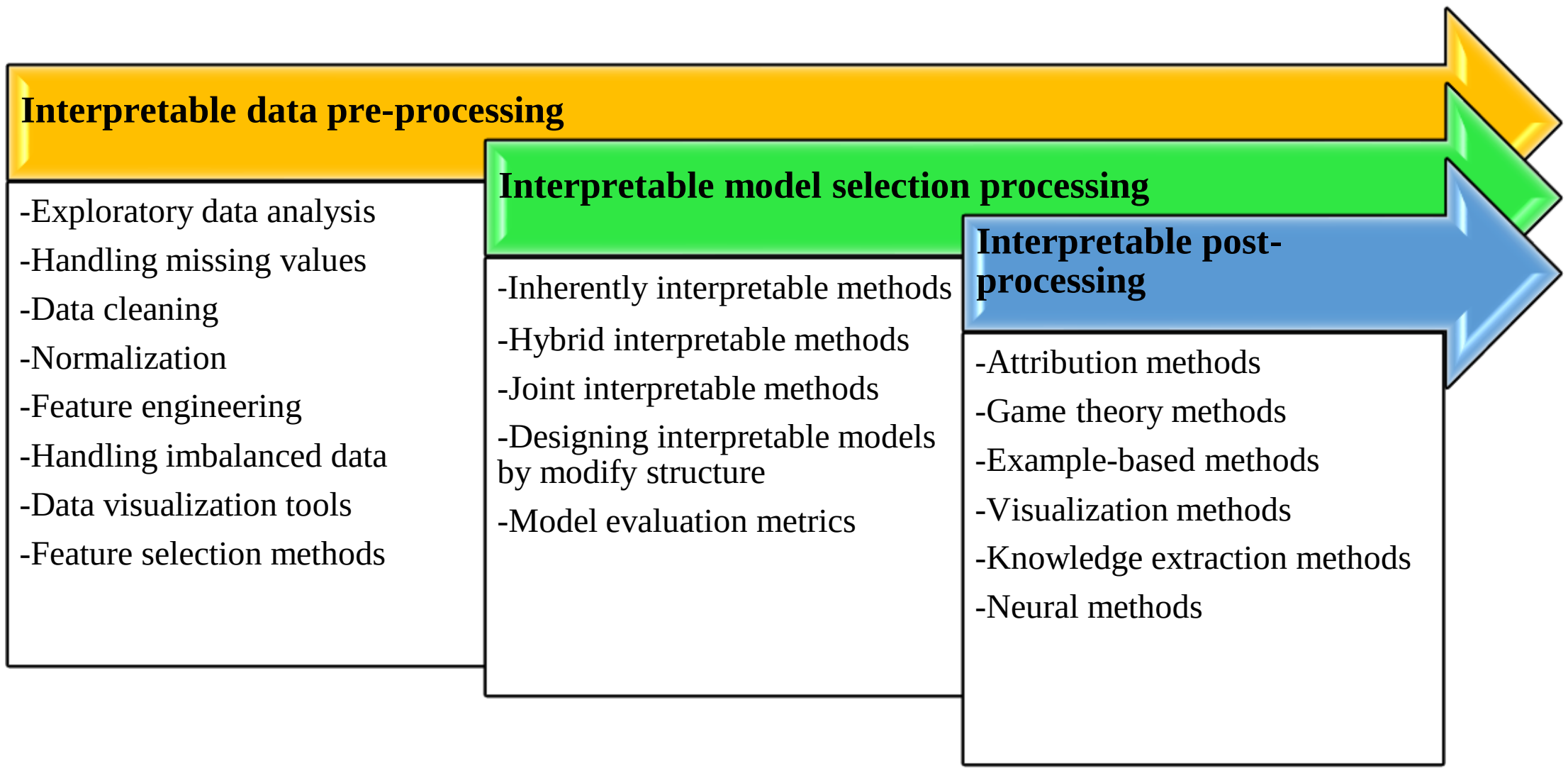

*Figure 9. Proposed 3-level interpretability process in intelligent health systems, detailing steps for interpretable data pre-processing, model selection processing, and post-processing, along with associated techniques and evaluation metrics.*

### *3.1. Data pre-processing interpretability and its health application*

The primary goal of this step is to understand and describe the data and its features used in model development. Interpretability pre-processing includes a wide range of tools that aim to better understand the dataset and the features used in training ML models. This level of explainability is critical due to the substantial influence of the training dataset on an AI model's decision-making process. Therefore, various interactive data analysis tools have been developed to aid in understanding the input data. Additionally, it is crucial to ensure that the collected data is cleaned, normalized, and balanced, as the model's performance relies on it. Understanding data is crucial for making AI systems more explainable, efficient, and robust. This involves techniques like exploratory data analysis (EDA), explainable feature engineering, standardizing dataset descriptions, using dataset summarization methods, and employing knowledge graphs. We will provide more details for each aspect of the pre-processing level and discuss related research in our included studies. Error! Reference source not found. shows a taxonomy of pre-processing explainability techniques.

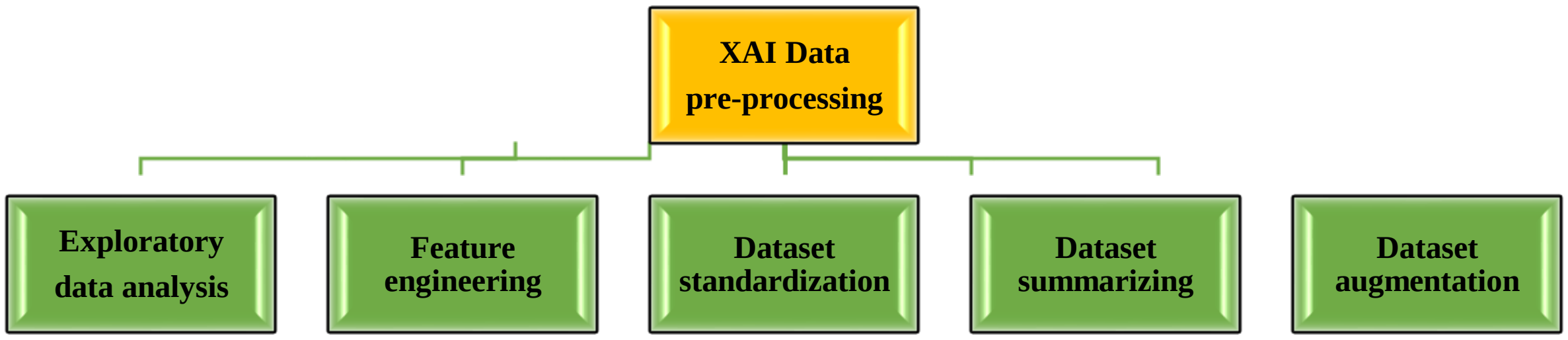

*Figure 10. Taxonomy of data pre-processing techniques for enhancing interpretability in XAI, including exploratory data analysis, feature engineering, dataset standardization, summarizing, and augmentation.*

#### *3.1.1. Considering diverse types of datasets*

The importance of conducting analysis based on specific data types cannot be overstated, particularly when dealing with high-dimensional or unstructured data. Different data types require tailored approaches for interpretation and understanding. For instance, high-dimensional data may cause dimensionality reduction techniques to extract meaningful insights effectively. Similarly, unstructured data, such as text or images, may require specialized methods like natural language processing or computer vision for comprehensive analysis. Understanding the intricacies of each data type allows for more effective interpretability techniques. Tailoring analysis methods to the specific characteristics of the data enhances the accuracy and reliability of interpretations, leading to more robust AI systems [45].

Furthermore, the adaptability of analysis techniques in response to the evolving nature of data types is vital for sustaining the relevance and efficacy of AI systems in a rapidly changing digital environment. As new forms of data emerge, such as sensor data from the Internet of Things (IoT) devices or streaming data from social media platforms, the strategies for data analysis and interpretation must also evolve. These emerging data types often blend characteristics of structured, unstructured, and semi-structured data, presenting unique challenges for data processing and analysis. For example, streaming data requires real-time analysis and decision-making capabilities. In contrast, sensor data from IoT devices demands robust algorithms capable of handling vast volumes of data in diverse formats.

In addition to considering diverse types of datasets, it is imperative to delve into interpretability per data type and explore scenarios where data may be multi-modal, needing fusion of interpretability results. This section aims to provide insights into the interpretability techniques tailored to various medical data types and the integration of interpretability results in multi-modal data settings.

***Medical note***. Interpretability techniques for medical notes, such as fact checking, often involve natural language processing (NLP) methods to extract meaningful information from textual data. Techniques such as sentiment analysis and topic modeling can offer valuable insights into patient narratives and clinical documentation [46].

***Clinical tabular data.*** Analysis of clinical tabular data may involve methods such as feature importance analysis and decision tree visualization to understand the relationships between different variables and their impact on clinical outcomes [47].

***Medical images.*** Interpretability techniques for medical images typically use computer vision approaches, including heatmaps and attention mechanisms, to highlight regions of interest and provide insights into diagnostic decisions made by AI models [48].

***Medical signals.*** Interpretation of medical signals, such as electroencephalogram (EEG) or (ECG) data, often requires time-frequency analysis and waveform visualization techniques to show patterns indicative of underlying physiological conditions [49].

Each data type presents unique challenges for interpretability. Medical notes may suffer from data sparsity and ambiguity, while images may show high dimensionality and variability. Signals may hold noise and artifacts that can affect interpretation accuracy. Understanding these challenges is essential for selecting proper interpretability techniques and ensuring the reliability of results.

***Multi-modal and multi-Centre data fusion.*** This paper [50] proposes using conceptual knowledge to train more explainable, robust, and less biased machine learning models, particularly in the medical domain where various modalities contribute to single results. The central question addressed is how to construct a multi-modal feature representation space spanning images, text, and genomics data using knowledge bases. Graph Neural Networks are advocated as a method to enable information fusion for multi-modal causality, with a focus on achieving a specified level of causal understanding. The paper aims to motivate the XAI community to delve into multi-modal embeddings and interactive explainability, emphasizing the pivotal role of Graph Neural Networks in helping causal links between features within graph structures.

#### *3.1.2. Exploratory data analysis (EDA)*

The goal of EDA is to gather important characteristics of a dataset, including its dimensionality, mean, standard deviation, range, and information on missing samples. Google Facets is a powerful tool for quickly extracting these features from the dataset [51]. Additionally, an EDA tool can find biases in the dataset, which may show a problem with class imbalance. When evaluating datasets, relying only on statistical characteristics is often insufficient. For instance, [52] proposed an explainable framework considering preprocessing by focusing on EEG data, applying an adaptive network to analyze data, providing details about their features, and implementing interpretable models like LIME and SHAP. To predict type 2 diabetes, in this paper [53], they applied exploratory data analysis, but they did not consider any feature engineering and visualization tools for the preprocessing part. Data visualization offers a range of charting options [54]; the most suitable type of chart depends on the dataset, application, and the specific statistical characteristics that a data scientist aims to communicate.

Real-world health and medical data are often complex and multidimensional, with many variables. In [55] and [72], due to the variety of shape sizes, boundary position and shape sizes, boundaries, positions, and shapes in medical image segmentation, a novel method has been proposed to enhance explainability in medical image segmentation. This method focuses on creating more understandable attention feature maps, allowing for the extraction of the most crucial features in distinct types of medical images, such as CT (Computed Tomography) and MRI (Magnetic Resonance Imaging) (Magnetic Resonance Imaging) scans. However, visualizing such high-dimensional medical data faces significant challenges. Using specialized charts like Parallel Coordinate Plots (PCP) is one approach to tackle this issue [56]. These charts help determine which features are important to keep or eliminate. Additionally, high-dimensional datasets can be transformed into lower-dimensional data. Principal Component Analysis (PCA) and t-distributed Stochastic Neighbor Embedding (t-SNE) are two well-known methods for this purpose [57], and the Embedding Projector toolbox supports the use of both techniques [58]. PCA is suitable when the underlying structure is mostly linear, while t-SNE is preferred otherwise. However, it is worth noting that t-SNE can be slow when applied to large datasets. In such cases, dimensionality reduction approaches like Uniform Manifold Approximation and Projection (UMAP) may be more efficient [59], which is claimed to be more exact and scalable compared to t-SNE.

#### 3.1.3. Dataset description standardization

Medical data are often shared without enough description. Standardization can help address problems like bias and missing values by allowing clear communication between data collectors and end-users. In addressing this challenge, tools like Datasheets for Datasets can be helpful for dataset description standardization [60], and Data Declarations for Natural Language Processing (NLP) [61]. These methods provide different frameworks for specific information related to a dataset and help in tracking the dataset's development, content, data collection process, and legal, and ethical considerations. Scholars have applied various methods for data standardization. For instance, in the domain of Myocarditis disease prediction using cardiac MRI, researchers in [62] implemented several steps for pre-modeling interpretability. They employed techniques such as noise removal, image resizing, as well as CutMix and MixUp algorithms for data augmentation. Additionally, in the study of human activity recognition [63], a creative method was introduced called XAI-HAR for feature extraction from data gathered by sensors placed at various locations within a smart home.

#### *3.1.4. Dataset summarizing methods*

Case-based reasoning is an interpretable modeling technique that predicts outcomes for a given input by comparing it to similar cases in the training data [64]. These similar cases, along with the model's predictions, can serve as explanations for the end-user. However, a significant challenge is the storage requirement for the entire training dataset, which can be costly or impractical for large datasets. To overcome this, a representative subset of the training data can be saved; dataset summarization aims to solve this problem. In data analysis, summarization techniques like document summarization [65], scene summarizing [66], and prototype selection [67] are used. Summarizing a dataset involves finding typical samples that give a quick overview of big data. For instance, in cardiovascular disease diagnosis with a PPG signal, in the data pre-processing stage, the data was segmented into 30-second intervals, and normalization was applied at the chunk level by subtracting the mean and dividing by the standard deviation [68]. Data squashing is another technique for data summarization [69], which aims to create a smaller version of a dataset that produces related results. Unlike data summarization, this method often assigns weights to samples in the smaller version of the dataset, using criteria similar to Bayesian learning.

#### *3.1.5. Data augmentation techniques*

When the number of samples is insufficient, efficient classification becomes challenging, and with a small size of the dataset, the risk of overfitting is significant. In such cases, the ML model struggles to learn from the data accurately and tends to memorize it instead. To address this issue, a widely used approach is data augmentation. Data augmentation involves generating artificial data points through various transformations applied to the existing dataset. In this context, SMOTE (Synthetic Minority Over-sampling Technique) can be viewed as a form of data augmentation. This is because, while balancing the dataset, SMOTE creates synthetic samples for the minority class, effectively expanding the overall size of the original dataset. SMOTE, introduced by [70], is an oversampling technique that relies on k-nearest neighbors to generate synthetic samples. For instance, in [71], a data augmentation method has been applied to prevent the negative effects of the small number of dental images for detecting caries. Moreover, Autoencoders (AE) techniques can be used to generate artificial data, as shown in this study: an AI framework for Early diagnosis of coronary artery disease, applying SMOTE, autoencoders, and CNN, resulting in higher accuracy[72].

#### *3.1.6. Explainable feature engineering*

Feature attribution refers to understanding the most notable features in the decision-making process of models [73]. The related features should also be easily explainable, and developers must understand their meaning and find the most relevant feature explanations for a specific end-user like health providers. In other words, the accuracy of predictive models is completely related to the attributes [74]. The two main approaches to feature engineering for interpretability are domain-specific methods, which rely on input from domain experts and insights from EDA, and model-based methods. Model-based feature engineering uses mathematical models to uncover the underlying structure of a dataset [75]. For example, in [76], a novel explainable method for the deep feature engineering model-based on Attention ResNet18 has been applied for asthma prediction. This process involved pre-processing, training the network, extracting deep features, prioritizing feature selection, and classification. Statistical analysis for data, pre-processing, and feature engineering have been used before model development in post-stroke stress and anxiety diagnosis [77]. The existing literature on pre-processing interpretability is not limited to the above paragraphs, the qualitative summary of works is shown in Appendix 5.

### *3.2.Interpretable process of model selection and its health application*

Even if the data is prepared, cleaned, and balanced carefully in the interpretability pre-processing stage, if the model does not have clear explainability, developers may still struggle to integrate their ability into the learning process to achieve improved results. Therefore, in addition to explainable pre-processing for data, understanding the model is critical; interpretable and exact ML algorithms are essential in applications such as medicine where errors can have severe consequences on people's lives. Interpretable processing tries to design models that are inherently interpretable and explainable. Using inherently explainable models for AI tool selection is often associated with adopting interpretable modeling. However, current methods still have a tradeoff between accuracy and interpretability. For instance, while decision trees are interpretable and widely applied in health and medicine for disease diagnosis, they are consistently surpassed in accuracy by less interpretable ML models like ensemble methods and DNN; despite their high computational and memory requirements [11]. The following sections will cover each aspect of processing interpretability and its application in healthcare and medicine. Also, you can find the comparison between the four aspects of interpretable modeling in Error! Reference source not found..

*Table 2. Comparison between four aspects of interpretable modeling.*

| | Definition | Advantage | Disadvantage |
|---|---|---|---|
| **Inherently** | Designed to be easily understandable by humans, with simple structures, White-Box models like DT and LR. | Clear and understandable by design.<br>Require no added explanation method.<br>Transparent relationship between features and predictions.<br>Well-suited for scenarios where interpretability is a critical requirement. | Limitations in handling complex relationships or high-dimensional data compared to more complex models.<br>Sacrifice predictive performance for interpretability in some cases. |
| **Hybrid** | Combine inherently interpretable with more complex models and aim to use the strengths of both | Achieve a balance between interpretability and accuracy. | It may require another effort to integrate and fine-tune both models. |
| | models: white-box and black-box. | Handle more complex relationships and high-dimensional data. | Interpretability might still be limited compared to inherently interpretable models. |
| **Joint** | Focus on generating and training models to provide explanations concurrently with predictions. | Provide interpretability linked to the predictions.<br>Generate explanations simultaneously during the prediction process. | Computationally intensive, cost-effective, and requires specialized architecture or techniques.<br>The quality of explanations varies based on the specific approach used. |
| **Modify Structure** | Modify the model's architecture to enhance its interpretability and make the model's decisions and behavior more understandable. | Does not necessarily require more explanation techniques or models.<br>It improves both prediction performance and interpretability? | Requires ability in model architecture and design.<br>Depending on the specific adjustments, there is still a trade-off between interpretability and accuracy. |

### *3.2.1. Inherently interpretable machine learning models (White-Box)*

The traditional approach to implementing interpretable models involves selecting White-Box methods like DT and LR. For instance, in [78], a DT method was designed as an interpretable ML technique to show hidden patterns in health records for dementia prediction. Additionally, for COVID-19 detection, an explainable gradient boosting framework based on DT was applied, which can be used with wearable sensor data [79]. Moreover, researchers in [80] developed a K-nearest neighbor (KNN) model for more explainable and easy-to-understand risk prediction of type-2 diabetes mellitus, hypertension, and dyslipidemia. For a better understanding and more correct stroke prediction models, researchers in [81] constructed Bayesian rule lists. However, selecting an interpretable model does not always guarantee explainability. Some models, like LR, can have difficulty simulating high-dimensional data, making them less explainable [23], to address this issue, some techniques, like regularization, can be used to simplify the model. Some researchers have shown that it is possible to create models that are both interpretable and high-performing. The challenge lies in finding the right balance between simplicity for understanding and complexity for accuracy. For instance, in [68], they have developed learned kernels to achieve interpretable and more exact PPG signal processing in cardiovascular disease detection.

### *3.2.2. Hybrid interpretable methods*

To create a high-performance and explainable model, one approach is to combine an inherently White-Box technique with a sophisticated Black-Box method [82]. This concept forms the basis of hybrid interpretable methods. On the other hand, the White-Box model can be used in the regularization step. Examples of such methods include deep k-nearest neighbors (DkNN) [83], deep weighted averaging classifier (DWAC) [84], self-explaining neural networks (SENN) [85], contextual explanation networks (CEN) [86], BagNets [87], Neural-symbolic (NeSy) models [88], and X-NeSyL (explainable Neural-Symbolic Learning) method [89]. In [62], the turbulence neural transformer (TNT) and explainable-based Grad Cam method were applied to the automatic diagnosis of myocarditis disease in cardiac MRI. A White-Box model (LR) was combined with a Black-Box model (ANN) for clinical risk prediction in acute coronary syndrome [90]. Also, to detect stroke in the early stage, [91] have applied a tree-boosting model and multilayer perceptron (MLPs), and provided a more explainable prediction.

#### *3.2.3. Joint prediction and interpretable methods*

Joint methods focus on simultaneously generating predictions and explanations. In other words, a complex model can be explicitly designed to clarify its decisions. For instance, the teaching explanations for decisions (TED) framework [92] is used to enhance training data with explanations for decisions. It combines output and explanations during training and provides user-friendly outcomes. Additionally, [93] have introduced a method similar to TED for generating multimodal explanations, which requires a training dataset with both textual and visual explanations. In [55], an explainable multi-module semantic guided attention-based network was designed for medical image segmentation, such as multi-organ CT images and brain tumor MRI images, which produced high performance between all metrics. Furthermore, neural additive models for understandable heart attack prediction were applied [94]. Interpretable predictive models with an attention-based recurrent neural network were applied for mortality prediction in the MIMIC-III database [95].

#### *3.2.4. Interpretability through architectural adjustments*

Modifying the architecture of a model can be a powerful way to enhance its explainability. For instance, [96] developed a CNN with an explainable structure using multiple EEG frequencies as input for abnormality prediction. The modified CNN achieved both more understandable information and higher accuracy. For brain tumor diagnosis, [97] designed a pre-trained vision language model (VLM) and proved its effectiveness in providing explainable predictions for medical applications. The existing literature on processing interpretability is not limited to the above paragraphs; the summary of included studies is shown in Appendix 5.

### *3.3.Post-processing interpretability and its health application*

In the first two sections of our overview of the Interpretability process, we investigated pre-processing interpretability and interpretable modelling, which focus on interpretability at the dataset level and during model development. However, these are comparatively smaller aspects of interest when compared to post-modeling explainability. This section, which concerns understanding after the model is built, is where the majority of XAI scientists have directed their focus and research efforts. The aim of the pre-processing interpretability is to provide user-friendly explanations to describe more understandable pre-developed methods. It can be categorized into six parts, and its application in healthcare will be discussed in the following paragraphs if applicable: (1) attribution methods, (2) visualization methods, (3) example-based explanation methods, (4) game theory methods, (5) knowledge extraction methods, and (6) neural methods [1].

#### *3.3.1. Attribution methods*

In medical image analysis, most attribution methods rely on pixel connections to show the significance of each pixel in the model's activation. This assigns a relevance or contribution value to each pixel in the input image [97]. Over the past few years, several new attribution techniques have appeared. There are four primary groups of attribution techniques: deep Taylor decomposition (DTD), perturbation methods, backpropagation methods, and Deep Lift. These will be summarized in Figure 11, but no other details about the specific method will be provided here as it has been thoroughly described in the literature. In [63], LIME with a RF classifier was used to develop an XAI-enabled human activity recognition model for those who have chronic impairments. Another study by [76] introduced an interpretable attention-based model for asthma detection using Gradient-weighted class activation mapping (Grad-CAM). This model proved both high performance and interpretability. In another study, LIME was used for activity recognition in elderly individuals using wearable sensors placed in various locations [19].

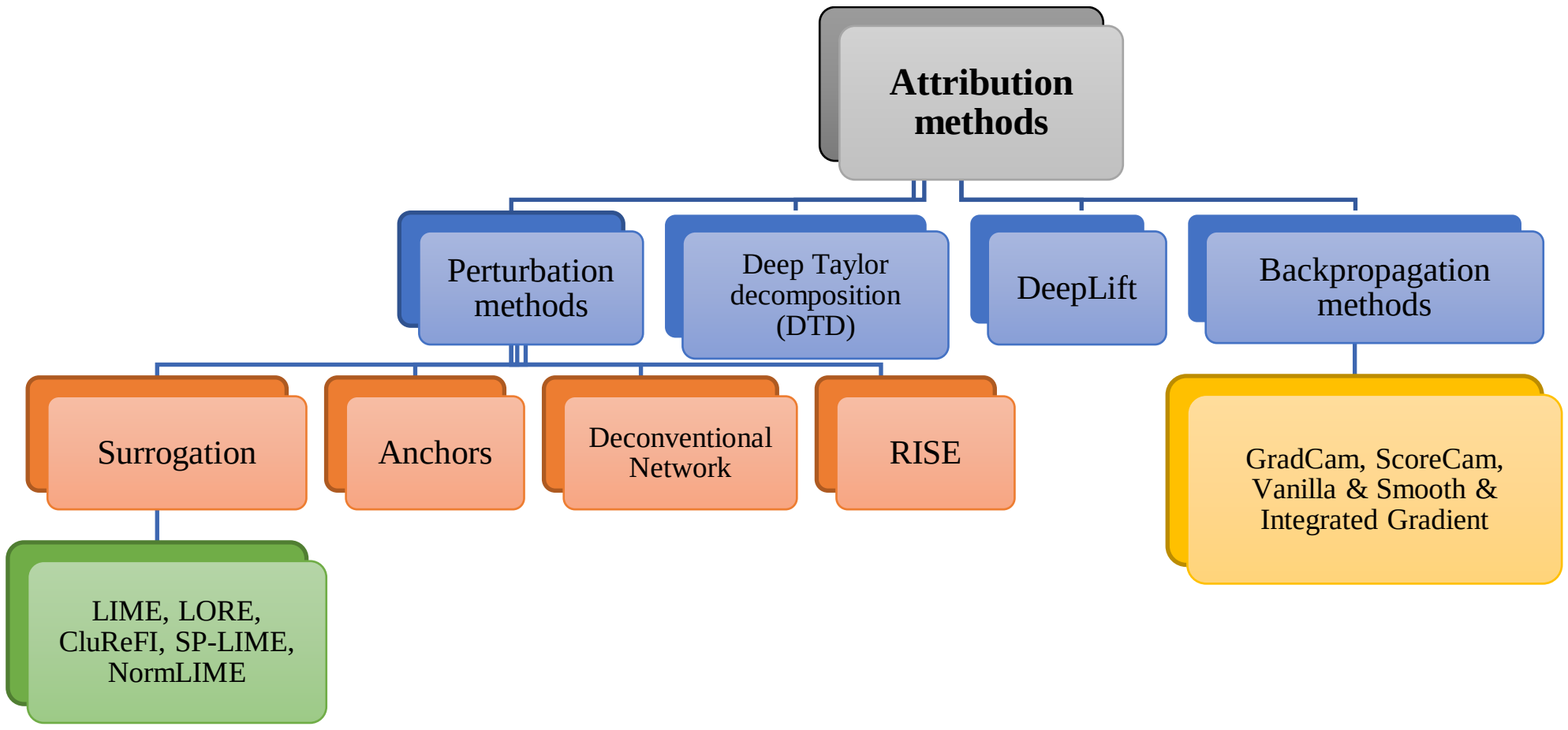

*Figure 11. Taxonomy of attribution techniques used in explainable AI, detailing various techniques such as perturbation methods, deep Taylor decomposition, and backpropagation methods, further breaking down into specific approaches like LIME, LORE, and Grad-CAM.*

### *3.3.2. Visualization methods*

Exploring the inner workings of an AI model by visualizing its representations to uncover hidden patterns is a fundamental idea. These visualization techniques are commonly applied to supervised learning models. Notably, this includes techniques like partial dependence plot (PDP) [98], accumulated local effects (ALE) [99], and individual conditional expectations (ICE) [100]. In the context of hepatitis and liver conditions, a transparent model was developed that integrates SHAP, LIME, and PDP. These tools were purpose-built to improve the clarity of decision-making [101]. Moreover, regarding the interpretability of ML models for predicting hypertension, this study [102] employed five global interpretability techniques (feature importance, partial dependence plot, individual conditional expectation, feature interaction, global surrogate models) and two local interpretability techniques (local surrogate models, Shapley value).

### *3.3.3. Example-based explanation methods*

Example-based explanations, also known as case-based explanations, include methods such as prototypes and criticisms [103], counterfactuals [104], and adversarial examples [105] for generating this type of explanation.

### *3.3.4. Game theory methods*

In 1953, Lloyd Shapley started exploring how each player in a game contributes. Later, this idea was applied in ML to understand the relationship between interpretability and predictions [106]. In this context, the 'game' refers to a single prediction in a dataset, the 'gain' is the difference between the actual prediction and the average of all predictions in the dataset, and the 'players' are the features that work together to figure out the gain. The Shapley value of a feature tells us how much it contributes to a specific prediction. Additionally, the SHAP provides a unified method for interpreting the output of any machine learning model. It uses the best Shapley values from coalition games to explain individual predictions [107]. The SHAP method is extensively used in healthcare applications. For example, in [18], an XAI model with SHAP was constructed to forecast the next day's perceived and physiological stress levels of pregnant mothers. This process involved finding crucial risk factors for predicting parental stress later. Furthermore, SHAP techniques were developed to predict the risk of amiodarone-induced thyroid dysfunction, serving as a tool for personalized risk assessment and aiding in clinical decision-making [17]. An interpretable model using the SHAP approach enhanced mental stress detection using heart rate variability data, providing detailed visualizations of outcomes for physicians and offering a deeper understanding of the results [108].

### *3.3.5. Knowledge extraction methods*

Understanding the internal workings of the Black-Box models is a challenge. For instance, in ANN algorithms, modifications to hidden layer filters/kernels can lead to complex internal representations. Extracting explanations from an ANN involves translating the knowledge bought by each layer into a format comprehensible to humans. The literature provides various techniques for extracting insights from Black-Box models, primarily relying on rule extraction [109] and model distillation methods [110]. For instance, in [108], a neural additive explanation was developed for heart attack prediction, enhancing the efficiency of model performance.

### *3.3.6. Neural methods*

Neural network interpretation techniques are vital for understanding and gaining insights from complex neural network models, making them more transparent and understandable. Methods such as feature importance and sensitivity analysis have been applied to interpreting neural networks [111]. Applications in healthcare have been presented for each aspect of pre-processing interpretability where applicable. However, it is important to note that the existing literature on post-processing interpretability extends beyond the methods discussed here. A summary of the studies included is provided in Appendix 5.

## 4. Application of interpretability in AI-based medical devices

In the realm of medicine, AI serves a dual purpose: it can complement existing medical devices or function independently as one. According to the European Medical Device Regulation, a medical device encompasses a wide array of instruments, software, implants, and materials designed for specific medical purposes, either alone or in combination [112]. AI-based medical devices form a category of health technologies that enhance human capabilities. They are used for predicting diseases, classifying data for disease management, perfecting medical treatments, and aiding in disease diagnosis. The U.S. FDA (Food and Drug Administration) (Food and Drug Administration) (Food and Drug Administration) (Food and Drug Administration) categorize AI-driven medical software as "Software as a Medical Device" when it is intended for disease prevention, diagnosis, treatment, or cure the increasing prevalence of digital health and AI-based medical devices underscores their potential to provide fair access to professional healthcare on a global scale [113]. This has the capacity to mitigate global health disparities and elevate the overall quality of life [114].

Despite the ongoing advancements in technologies such as IoT-based medical sensors and wearables, telemedicine, medical LLMs, and digital care twins, concerns persist among end-users, including patients and clinicians, about the trustworthiness of these technologies and their outcomes. Addressing these concerns needs answering several critical questions: How can we enhance the interpretability of digital health technologies? How can we effectively communicate deliverable outcomes to end-users? How can we improve interpretable decision-making for clinician-AI communication in AI-based medical devices? How can we ensure diversity and inclusion in an intelligent health system? [26, 115].

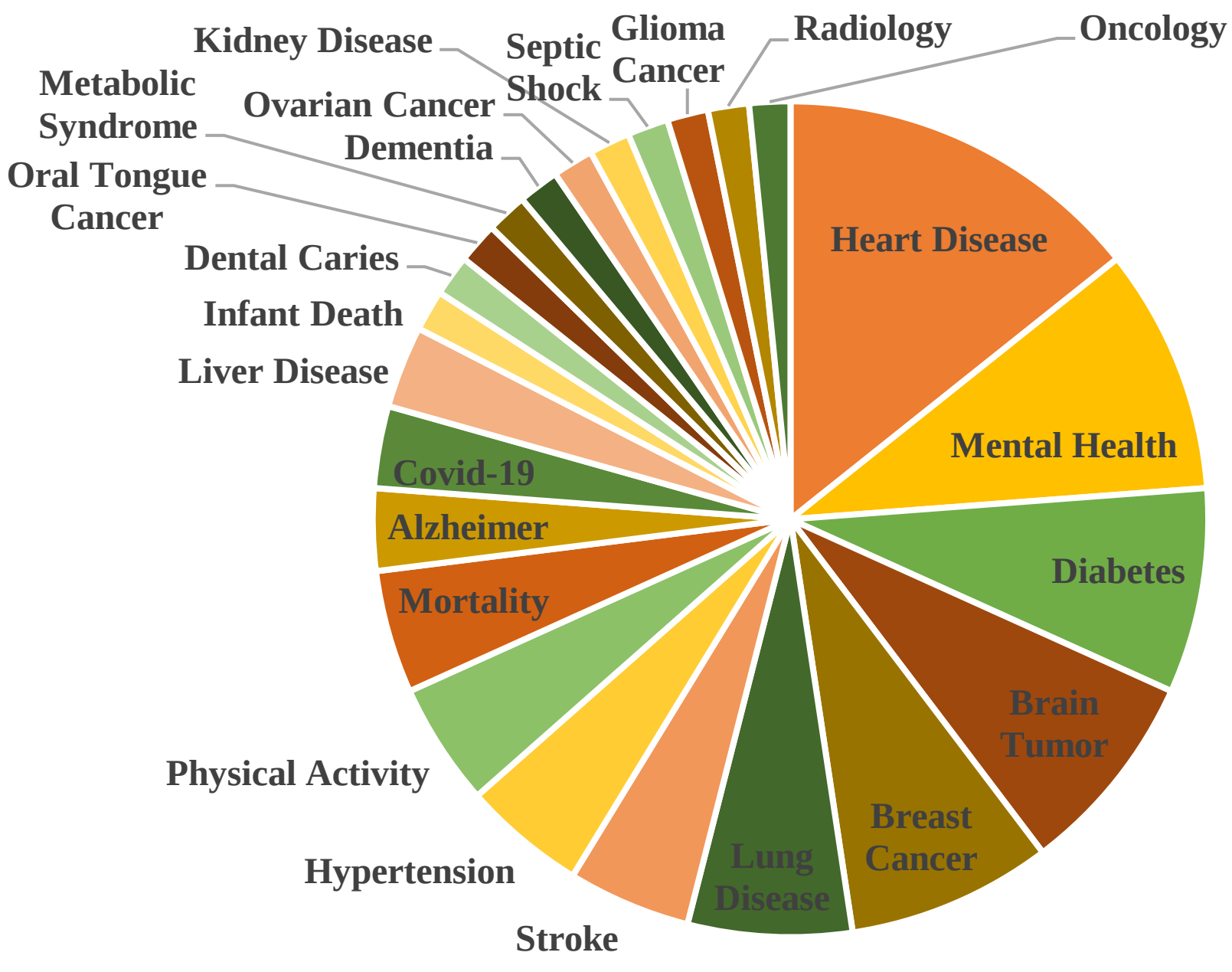

*Figure 12. Comparison between current XAI applications in AI-based MDs illustrates a pie chart comparing the distribution of current XAI applications in AI-based medical diagnostics, highlighting a variety of conditions such as Covid-19, Alzheimer's, cancer types, and other diseases, highlighting the diverse areas where XAI is used.*

In this section, we conducted searches using phrases such as telemedicine, mHealth, digital care twins, medical LLMs, biosensors, and wearables. However, we only found applications of XAI in the next section, where we will delve into interpretability and its application in medical sensors, wearable health trackers, and medical large language models. To the best of our knowledge, this systematic review stands for the first comprehensive examination of XAI applications covering these categories of digital health technologies. Error! Reference source not found. offers a comparison of current XAI applications in AI-based MDs.

### *4.1.Interpretability applications in IoT-based biosensors and wearable medical devices*

The integration of body sensor data into healthcare has captured considerable attention from researchers owing to its diverse applications in smart healthcare systems [116, 117]. However, this integration has presented challenges, particularly concerning transparency and explainability. XAI addresses these concerns by offering users comprehensive insights into the inner workings of the model [118].

By integrating XAI into the Internet of Things (IoT)-based medical sensors and wearable health trackers, we can develop a system that efficiently manages Big Data and enhances transparency and interpretability for end-users. This integration empowers users to understand better and trust the insights provided by these devices, fostering more informed decision-making in healthcare settings. Table 3 provides a summary of existing XAI applications in wearable medical devices (WMD). However, the use of XAI in the Internet of Medical Things (IoMT)-based medical devices extends beyond the scope of Table 3, which presents only a choice of notable examples in this domain.

*Table 3. Examples of wearable health tracker applications employing interpretability processes.*

| Ref | Application | IoT Wearable Medical Devices |
|---|---|---|
| [119] | Diabetes in pregnant women | Blood pressure and pulse rate sensors |
| [20] | Respiratory infections - COVID-19 | HRV and BPM with wrist-worn biometric wearables |
| [120] | Respiratory symptoms for the pandemic | MS Band 2 wristband – monitor cough and fever |
| [2] | Hypoglycemia detection and diabetes | Smartwatch sensors - detect hypoglycemia |
| [63] | Human activity recognition | Motion sensors, temperature-sensitive devices |
| [121] | Chronic obstructive pulmonary disease (COPD) monitoring | Smartphone, ECG patch providing day/night movement monitoring, pulse meter providing oximetry monitoring, weight scale, sphygmomanometer for blood pressure monitoring. |
| [122] | Recognizing the physical activity of older people | Wearable wellness system (WWS), wearable WBan system |
| [19] | Human activity recognition | ECG electrodes, accelerometer, gyroscope, magnetometer |
| [123] | Mental stress detection | Wrist sensors (EDA data) and chest-worn (HRV data) |
| [68] | Monitor various cardiovascular parameters. | Chest PPG, ECG, accelerometer, electrodermal, and wrist PPG |

### *4.2.Interpretability applications in ChatGPT and medical LLMs*

Recent advancements in AI have led to the development of sophisticated Large Language Models (LLMs) such as GPT-4 and Bard. These models hold significant promise in healthcare, offering potential applications ranging from aiding in clinical documentation to serving as chatbots for patient inquiries. However, of LLMs into healthcare requires careful consideration due to their unique characteristics, which sets them apart from regulated AI technologies, particularly in the critical context of patient care [124]. Regulating LLMs, including models like GPT-4 and other generative AI systems, in healthcare poses a challenge. It requires striking a delicate balance between ensuring safety, upholding ethical standards, and safeguarding patient privacy, all while using transformative potential. Without adequate human oversight and responsible implementation practices, applications of generative AI in healthcare risk spreading misinformation or generating inaccurate content, potentially compromising patient care and trust in the healthcare system. Therefore, robust regulatory frameworks and ethical guidelines are essential to guide the integration and deployment of LLMs in healthcare settings, ensuring that they help patients while upholding the highest standards of safety, accuracy, and privacy protection.

It is important to note that these tools are known to have the potential for errors, misinformation, and biases. For instance, ChatGPT lacks transparency due to its opaque nature, which may pose a challenge for healthcare professionals who require clear explanations. Additionally, biases in the training data may change accuracy, potentially leading to incorrect diagnoses or treatment recommendations [125]Therefore, medical professionals must thoroughly review and validate ChatGPT suggestions before incorporating them into the clinical decision-making process, from pre-processing to post-processing. Recently, researchers have used interpretability and explainability techniques in applying LLMs in the medical domain. Error! Not a valid bookmark self-reference. highlights a selection of notable examples in this field, although only a few studies have applied XAI in LLMs. Additionally, detailed descriptions of specific methods are not provided here as they have been thoroughly described in existing literature.

*Table 4. List of medical large language models using interpretability process.*

| Ref | Application | Medical LLMs and ChatGPT |
|---|---|---|
| [126] | Mental Health Care | Pre-trained language models (PLMs), ChatGPT, Prompt Engineering |
| [97] | Medical Image Diagnosis | Pretrained vision-language models (VLMs) such as CLIP |
| [111] | Heart Disease | Contextual prompts, zero-shot and few-shot prompt learning based on LLMs |

## 5. Quality evaluation and improvement of the interpretability process

In this section, we delve into assessing and evaluating the effectiveness of our proposed interpretability framework for healthcare and medicine. Our primary focus is evaluating the comprehensibility of the explanations by end-users, who often serve as the ultimate decision-makers. We specifically prioritize individuals holding key roles in healthcare, such as patients, doctors, nurses, caregivers, health system managers, and other domain experts.

This evaluation is crucial for gauging the utility and efficacy of our explainable artificial intelligence (XAI) system in practical, real-world scenarios. This segment encompasses (1) accuracy, (2) reliability, (3) robustness, (4) interpretability, (5) usability, (6) human-AI interaction, (7) ethical considerations, (8) responsiveness to feedback, (9) compliance with regulatory standards, (10) clinical validation and evidence base, (11) transparency, and (12) scalability [192]. Table 5 offers a comprehensive description of all these aspects, while Error! Reference source not found. provides a visual representation of the quality level for the explainability process.

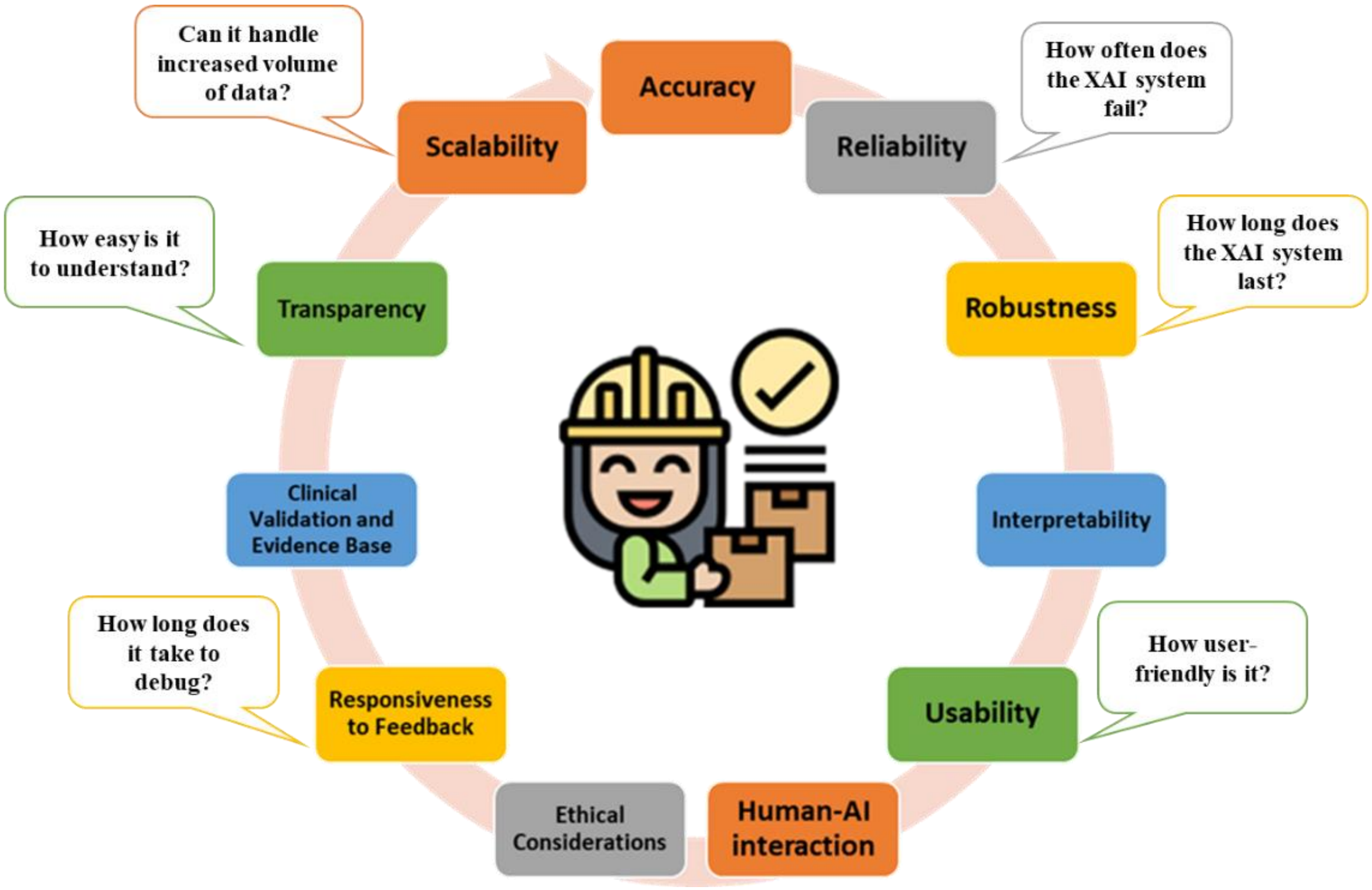

*Figure 13. Illustration of interpretability quality assessment in the context of human-AI interaction, mapping key attributes like scalability, reliability, robustness, usability, and transparency.*

*Table 5. Description of XAI system quality control and improvement dimensions.*

| Dimension of quality in XAI system | Description of each dimension in healthcare applications |
|---|---|
| **Accuracy** | Will the XAI system provide correct and reliable outputs, diagnoses, or recommendations? |
| **Reliability** | How often does the XAI system fail? (The consistency of XAI systems in producing exact results across different scenarios and datasets.) |
| **Robustness** | How long does the XAI system last? (The ability of XAI systems to perform effectively and keep accuracy even in the presence of noisy or incomplete data.) |
| **Interpretability** | How is it easy to understand? (The extent to which the XAI system's decision-making process can be understood and explained to users.) |
| **Usability** | How is it user-friendly and accessible for healthcare professionals to interact with and integrate into their workflows? |
| **Human-AI interaction** | How is it easy to be incorporated into existing healthcare IT (Information Technology) infrastructure and workflows? |
| **Ethical considerations** | XAI systems must run in a manner consistent with ethical guidelines, respect patient privacy, and avoid biases. |
| **Responsiveness to feedback** | How long does it take to debug and provide a request? Is it helpful? (The XAI system's ability to learn and adapt based on feedback from healthcare professionals and patients.) |
| **Compliance with regulatory standards** | Adherence to legal and regulatory requirements governing the use of XAI in healthcare. |
| **Clinical validation and evidence-based** | There is a wealth of rigorous clinical studies and evidence proving the effectiveness and reliability of the XAI system in real-world healthcare settings. |
| **Transparency** | The openness and clarity with which the XAI system's capabilities, limitations, and potential risks are communicated to users. |
| **Scalability** | The ability of the XAI system to handle increased volumes of data and users without a significant loss of performance. |

## 6. Step-by-step roadmap to implement responsible clinician-AI-collaboration framework

This section outlines a systematic approach for integrating robust explainability and interpretability into healthcare and medicine, as depicted in Error! Reference source not found.. A crucial part of the successful implementation of robust XAI and IML processes within health systems involves setting up a Health-to-Data Center (D2H). This center encompasses four essential components: (1) Clinician-AI Communication Desk, (2) Data Scientist Help Desk, (3) Quality Assessment Board, and (4) Ethical Review Board. The center emphasizes close collaboration among these groups to help step-by-step decision-making with AI in healthcare. Error! Reference source not found. offers detailed insights for implementing this framework in intelligent health systems, encompassing pre-processing, processing, and post-processing stages, followed by evaluation of the process. It explains the ML decision tailored to meet specific end-user needs.

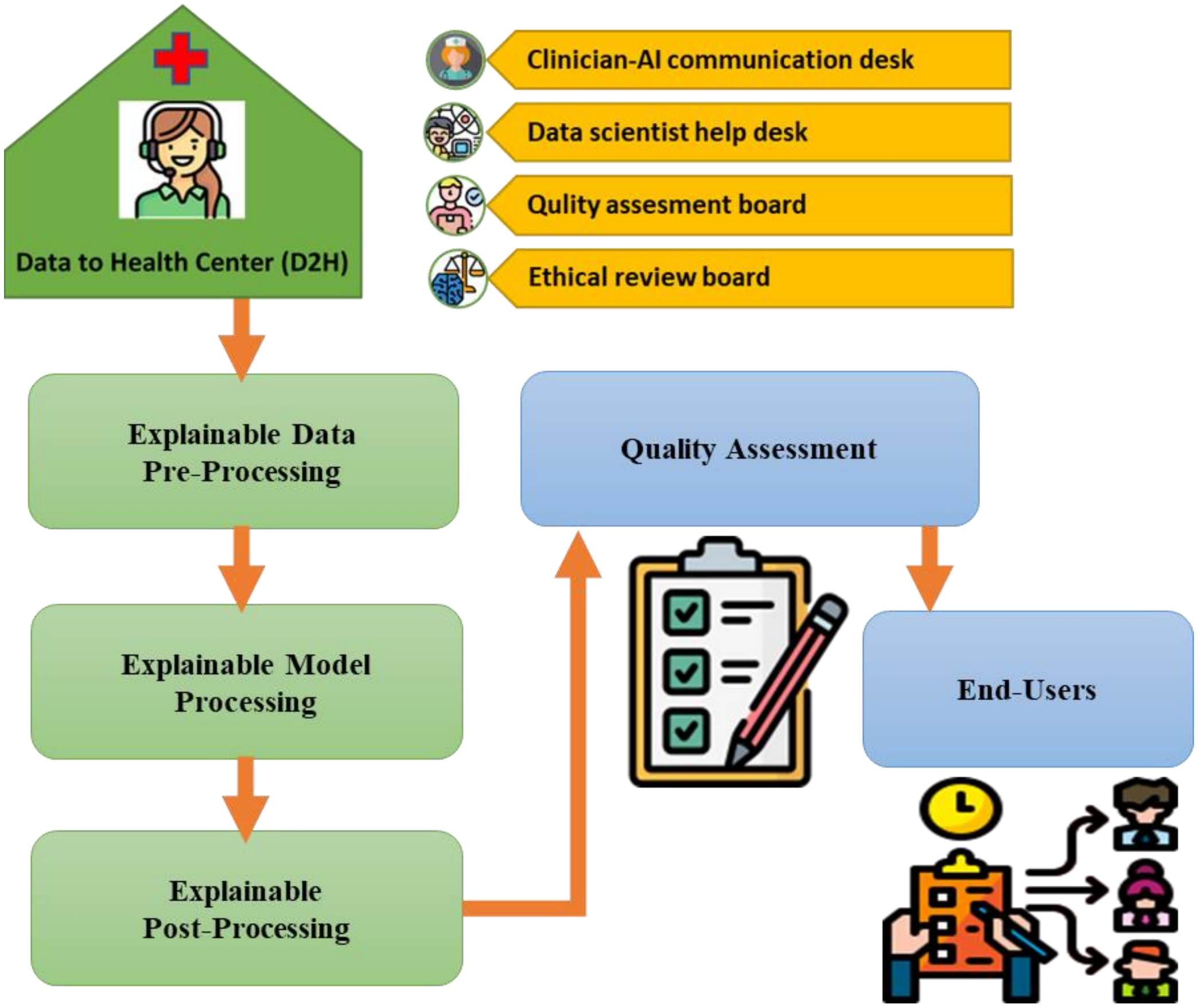

*Figure 14. Responsible clinician-AI-collaboration framework to implement trustworthy AI in healthcare, showing the flow from a Data to Health Center through stages of explainable data pre-processing, model processing, and post-processing, culminating in quality assessment and interaction with end-users.*

In the entire process of the three-level interpretability approach, from data pre-processing to model selection and post-processing, it is crucial to consider the questions provided in Figure 9. By contemplating these questions, we can offer more comprehensive explanations for each step of our AI systems tailored to specific end-users, such as patients, clinicians, health managers, or developers. Following the explainability process, it is critical to develop and design quality assessment tools to evaluate the decisions made by ML methods and the robustness of the XAI system. In the context of robustness, a comprehensive evaluation of the ML decision-making process includes assessing the resilience and stability of the models [127]. Robustness is ensured by incorporating methodologies that address potential adversarial attacks, data perturbations, and variations in real-world scenarios. Additionally, it is crucial to design and implement quality assessment tools that evaluate the accuracy and efficacy of the ML decision and gauge the robustness of the entire XAI system. These tools play a critical role in confirming the system's reliability under diverse conditions and contribute to building trust in the decision-making capabilities of AI within healthcare settings [128].

## 7. Discussion

AI-based medical devices have become increasingly prevalent in healthcare and medicine, as shown by the distribution of published studies applying IML and XAI in medical settings over the years. This trend underscores the growing importance of developing a responsible and interpretable AI framework to enhance trust among end-users in health systems, particularly between clinicians and AI. Appendix 5 provides a comprehensive summary of the papers examined in this review, shedding light on the methods and approaches used in many studies.

According to the review study, SHAP and LIME are among the most often employed XAI methods in healthcare applications. Post-processing methods have appeared as the predominant approach for providing explainability in health applications, as revealed by our analysis. Interestingly, the overview of examined XAI methods shows that black-box models are more commonly used than white-box methods for developing interpretable systems for clinical decision support. Despite employing complex techniques to achieve high accuracy, researchers often see post-hoc explainability methods to elucidate their intricate frameworks and enhance interpretability.

Our findings underscore a notable gap in the application of XAI and IML methods to wearable medical devices and biosensors for health tracking. Despite the increasing use of these devices, only a small fraction of studies in this area have utilized XAI methods to enhance the interpretability of decisions made by wearable health trackers. Similarly, there is limited research applying XAI methods to LLMs and generative AI for healthcare applications, showing a potential area for further exploration.

Among the 74 articles reviewed, only around 10 experimental results are presented in The integration of body sensor data into healthcare has captured considerable attention from researchers owing to its diverse applications in smart healthcare systems [116, 117]. However, this integration has presented challenges, particularly concerning transparency and explainability. XAI addresses these concerns by offering users comprehensive insights into the inner workings of the model [118].

By integrating XAI into the Internet of Things (IoT)-based medical sensors and wearable health trackers, we can develop a system that efficiently manages Big Data and enhances transparency and interpretability for end-users. This integration empowers users to understand better and trust the insights provided by these devices, fostering more informed decision-making in healthcare settings. Table 3 provides a summary of existing XAI applications in wearable medical devices (WMD). However, the use of XAI in the Internet of Medical Things (IoMT)-based medical devices extends beyond the scope of Table 3, which presents only a choice of notable examples in this domain.

Table 3, proving the application of XAI methods in wearable health trackers. Additionally, only three studies in Recent advancements in AI have led to the development of sophisticated Large Language Models (LLMs) such as GPT-4 and Bard. These models hold significant promise in healthcare, offering potential applications ranging from aiding in clinical documentation to serving as chatbots for patient inquiries. However, of LLMs into healthcare requires careful consideration due to their unique characteristics, which sets them apart from regulated AI technologies, particularly in the critical context of patient care [124]. Regulating LLMs, including models like GPT-4 and other generative AI systems, in healthcare poses a challenge. It requires striking a delicate balance between ensuring safety, upholding ethical standards, and safeguarding patient privacy, all while using transformative potential. Without adequate human oversight and responsible implementation practices, applications of generative AI in healthcare risk spreading misinformation or generating inaccurate content, potentially compromising patient care and trust in the healthcare system. Therefore, robust regulatory frameworks and ethical guidelines are essential to guide the integration and deployment of LLMs in healthcare settings, ensuring that they help patients while upholding the highest standards of safety, accuracy, and privacy protection.

It is important to note that these tools are known to have the potential for errors, misinformation, and biases. For instance, ChatGPT lacks transparency due to its opaque nature, which may pose a challenge for healthcare professionals who require clear explanations. Additionally, biases in the training data may change accuracy, potentially leading to incorrect diagnoses or treatment recommendations [125]Therefore, medical professionals must thoroughly review and validate ChatGPT suggestions before incorporating them into the clinical decision-making process, from pre-processing to post-processing. Recently, researchers have used interpretability and explainability techniques in applying LLMs in the medical domain. Error! Not a valid bookmark self-reference. highlights a selection of notable examples in this field, although only a few studies have applied XAI in LLMs. Additionally, detailed descriptions of specific methods are not provided here as they have been thoroughly described in existing literature.

Table 4 have explored using XAI methods in LLMs and ChatGPT for healthcare purposes. These findings underscore the need for more research in these areas to improve the trustworthiness and interpretability of AI-driven healthcare systems.

Our study emphasizes the critical role of interpretable processes in enhancing trust in clinical decision-support systems and providing easily understandable results for specific end-users. We propose the establishment of a data-to-health center, fostering close collaboration among clinicians, data scientists, and ethical and quality control boards to ensure the robustness and trustworthiness of health system performance. Figure 15 provides a comprehensive summary of the studies included in our review, highlighting key findings and areas for future research.

The pie chart in the Figure 15 stands for the distribution of several types of AI tools used in the medical studies reviewed. It shows that a vast majority (80%) of the tools are classified as AI tools. This category encompasses a broad range of AI technologies. Meanwhile, 15% of the tools are Wearable Medical Devices (WMD), showing a moderate representation in the studies. The smallest part (5%) is attributed to LLMs like ChatGPT, suggesting their emerging but still minor role in the context of the reviewed studies.

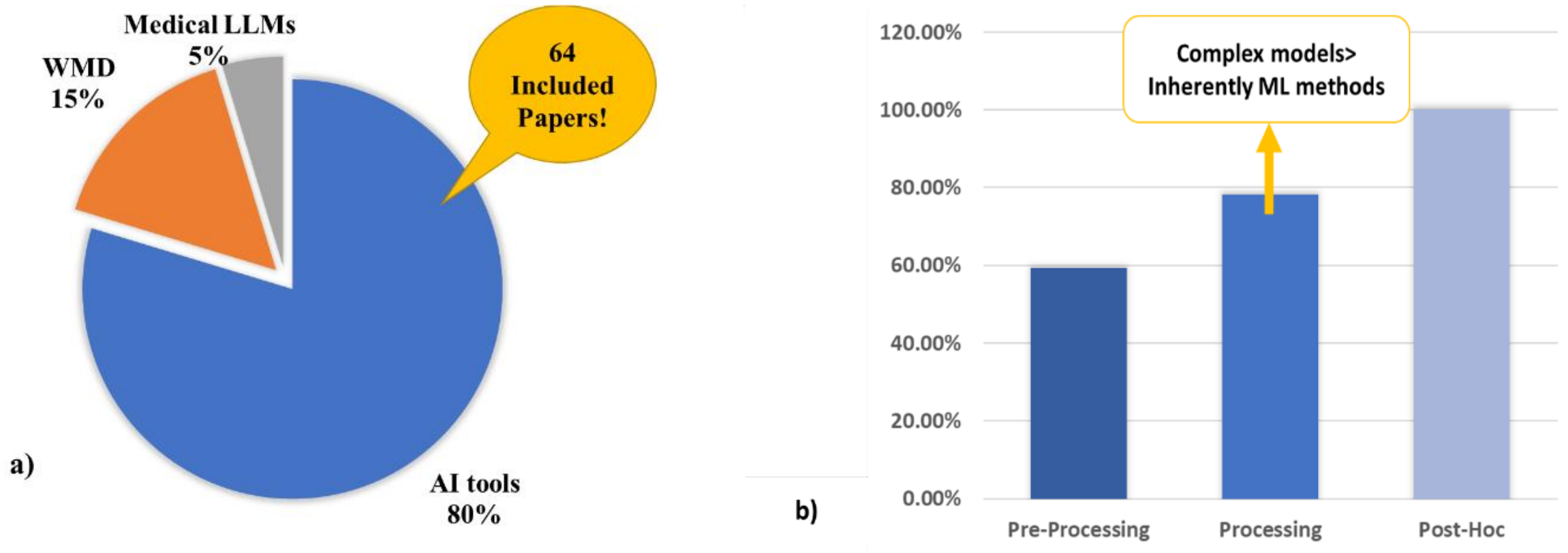

*Figure 15. Summary of two-part graphical summary of included articles: (a) a pie chart distribution of AI-based medical devices (MDs) showing the predominance of AI tools, and (b) a bar chart displaying the distribution of machine learning process stages, namely pre-processing processing, and post-hoc analysis.*

The bar graph illustrates the prevalence of different approaches to explainability at various stages of processing within healthcare applications. The stages are pre-processing, processing, and post-hoc. The data shows that post-hoc methods are used in over 100% of the cases, which suggests that in some studies, more than one post-hoc method may have been applied. The processing stage is also significant, though less than post-hoc while pre-processing shows the least use among the three. This trend shows a preference for explaining AI decisions after the fact (post-hoc), rather than integrating explainability into the model itself (pre-processing) or during the model's operation (processing).

Figure 15 supports the narrative that although AI tools are widely used in medical studies, the specific areas of wearable devices and language models are less explored. Additionally, the preference for post-hoc explainability suggests a trend in the field towards providing explanations after AI systems have made their decisions, which may reflect the complexity of integrating explainability into more dynamic, real-time systems.

## 8. Open challenges and future insights for XAI in health systems

It is important to highlight that despite the abundance of XAI strategies, metrics, and tools, many questions remain. For instance, which method provides the most effective explanations, and how should we evaluate the quality of these explanations? How can we tailor explanations to a specific end-user? How can we strike a balance between performance and interpretability? How can we design our system for specific health applications and ensure continuous improvement? This section examines the challenges of implementing the interpretability process in healthcare and medicine. We discuss concerns in this field and show what must be achieved to enhance trust among end-users in clinical decision support systems.

### *8.1. Misuse of post-processing explainers in health application*

The current Black-Box nature of AI has raised concerns about its use in critical areas such as medicine. There is a belief that explainable AI could build trust among healthcare professionals, offer transparency in decision-making, and potentially help address biases. However, we argue that these expectations for explainable AI may not be realistic due to the intrinsic issues with post-hoc explainers that may make their results untrustworthy. It is unlikely that current methods will fully achieve the goals of providing decision support at the patient level. The paper posits that relying solely on post-processing explanation algorithms to meet the legal requirements for explaining machine learning algorithms in healthcare may not be effective. This ineffectiveness stems from the need for explanations in situations where various parties have conflicting interests, wherein the explanation might be manipulated to serve a particular agenda. The inherent uncertainty of these explanations further complicates the issue, making it difficult to meet the transparency goals mandated by law. The paper recommends that there be a more open and honest discussion about the potential and limitations of post-processing explanations, especially in clinical decision support systems, which directly affect patients' lives and where conflicting interests may arise.

### *8.2.Designing XAI framework for specific health application*

This section underscores the importance of adopting a structured roadmap for the application of XAI in healthcare and medicine, an approach embodied by what is termed the "Data to Health Center." As depicted in Error! Reference source not found.4, this framework significantly focuses on the pivotal role of end-users. It also brings attention to the necessity of tailoring the XAI framework to meet the unique requirements of each specific health application or goal, ensuring that relevant queries are considered at every stage of the process.

A crucial aspect of this tailored approach is the ongoing effort to balance performance with interpretability. This involves a delicate compromise between developing high-performing AI models and keeping a level of transparency that is understandable to users, thereby fostering trust and ensuring that the AI's decision-making processes can be assessed and confirmed. Finding this balance is essential, as it directly changes the effectiveness and acceptance of AI in clinical settings. The roadmap implies an iterative, user-centric design process where performance metrics and interpretability guidelines are continuously refined in collaboration with healthcare professionals to meet the evolving demands of patient care and medical research [129].

### *8.3. Trade-off between interpretability and accuracy*

The discussion on the trade-off between interpretability and accuracy addresses a fundamental aspect of XAI in healthcare systems. It has a profound impact on how end-users, including patients and healthcare professionals, perceive and trust the decisions made by ML models. The concept posits that as the explainability of an AI system increases, its performance might inversely decrease. As a result, AI systems are often categorized into Black-Box, Grey-Box, and White-Box models, depending on the level of transparency they offer in their decision-making processes [130-132].

Ideally, healthcare systems receive help from models that provide both important levels of explainability and satisfactory performance. Nevertheless, a compromise often arises between the system's ability to present patterns in a manner understandable to humans and its ability to fit complex data accurately. This balance is critical as it involves mitigating the clinical and human risks that come with the potential for misclassification, an issue that should be clearly communicated to end users to align their expectations and understanding of the system's capabilities and limitations [21]. Furthermore, achieving a high degree of transparency is not limited to the models alone. It also encompasses data pre-processing—ensuring that the data fed into the model is processed in an explainable manner, which significantly affects the overall interpretability of the ML model. This implies that for an AI system in healthcare to be fully interpretable, the entire pipeline, from data preprocessing to model decision-making, must be designed with transparency in mind.

### *8.4.Ethical guidelines (Inclusion and diversity)*

The section on Ethical guidelines (Inclusion and diversity) highlights the evolution of principles and guidelines for ethical AI by various organizations amidst ongoing debates about the definition of ethical AI and the standards necessary for its application. The text underscores the critical yet often overlooked aspect of gender equity in digital health initiatives, particularly worsened during the pandemic. These initiatives, while aimed at empowering women by improving healthcare access and reducing unpaid care work, have inadvertently contributed to gender disparities. This is due to issues like increased domestic violence, higher female unemployment, and more burdens of familial care, fueled by limited access and prevailing harmful stereotypes. This situation underscores the urgent need for dedicated efforts to address gender inequities within digital health contexts [26].

Furthermore, the application of ethics in AI, especially within healthcare, is presented as a multifaceted challenge that requires interdisciplinary collaboration. The spectrum of ethical concerns includes fairness, bias, privacy, and security, which can be technically mitigated through strategies such as data preprocessing and algorithmic adjustments. However, ethical considerations are also deeply influenced by the specific application domain. For instance, medical AI systems bring forth unique ethical challenges related to patient safety and privacy, causing specialized technical and legal frameworks [133] [134].

The navigation of ethics in AI is described as requiring a bespoke, domain-specific approach. Initiatives like the IEEE Global Initiative for Ethical Considerations in AI and Autonomous Systems and the EU Ethics Guidelines for Trustworthy AI are highlighted for their contributions to setting up frameworks and recommendations for the ethical development and implementation of AI. These efforts reflect a growing recognition of the need for rigorous ethical guidelines that are sensitive to the nuances of various application domains, including healthcare, to ensure that AI technologies are developed and deployed in a manner that is fair, secure, and respects the privacy and safety of all individuals [135] [136].

### *8.5.User-Centered AI system*

This section emphasizes the significance of tailoring AI systems in healthcare to the needs and understanding of end-users, including patients and clinicians. It notes the importance of demystifying the internal workings and outcomes of ML methods to foster trust within these key stakeholder groups. Highlighting research findings, it points out that current efforts in interpreting and explaining AI decisions are developer-centric, often overlooking the actual needs of end-users. This misalignment calls for a more user-focused validation of AI systems to ensure they meet the benchmarks of transparency, accountability, and fairness [137-140].

The discourse further underlines the necessity of close collaboration between ML experts and clinicians to bridge the communication gap between AI technologies and clinical practice. Such collaboration is pivotal not just for enhancing the mutual understanding between these groups but also for refining AI systems based on clinical insights. Clinicians' critical role in interpreting XAI outputs and pinpointing opportunities for model enhancement is particularly stressed [141] [142].

Moreover, the section touches on the different expectations and preferences for explanations among specialists and the general user base. While ML professionals might gravitate towards more technical, mathematical explanations, clinicians and patients may find visual or more intuitively understandable explanations more useful. This discrepancy underscores a broader challenge in the field of XAI—achieving a balance between technical accuracy and user comprehensibility in explanations [143]. Sociological research is cited to suggest that effective explanations are those that are concise, socially acceptable, and capable of contrasting different outcomes or scenarios, thereby pointing towards a need for designing XAI outputs that cater to a diverse audience with varying levels of technical ability. Addressing these differences in preferences and feelings is crucial for the successful implementation of XAI in healthcare, ensuring that AI-driven insights are accessible, actionable, and meaningful to all end-users [144] [145].

### *8.6.Evaluation metrics (Quality control tools)*

When assessing the interpretability of AI in healthcare systems, it is essential to use the right evaluation metrics. The complexity of applying uniform quality control metrics across differing goals presents a significant challenge for XAI systems. A practical solution involves adopting various scales tailored to the specific dimensions of each evaluation, enabling a nuanced differentiation in the assessment process. For example, in gauging user trust, a range of variables can be meticulously analyzed through different scales, employing methodologies such as surveys and interviews. This allows for a comprehensive understanding of trust as it relates to the AI system. Furthermore, to effectively evaluate the quality of explanations provided by AI systems, it is important to consider user satisfaction concerning the explanations' comprehensibility, usefulness, and adequacy of the information provided. These factors can be assessed through targeted feedback mechanisms designed to capture the user's experience and belief of the AI system [146].

In the context of iterative design, a strategic balance of diverse design methodologies and assessment types is crucial. This ensures that the design goals of the AI system are in harmony with the evaluation metrics chosen, particularly for healthcare applications. Such an approach not only aids in refining the AI system to meet specific healthcare needs but also aligns with the overarching goal of making AI systems more transparent, understandable, and user-friendly for healthcare professionals and patients alike. This alignment is key to fostering trust and confidence in AI-driven healthcare solutions, contributing to better health outcomes and more efficient care delivery [147] [148].

### *8.7. Cognitive and usability engineering*

Cognitive engineering in healthcare is centered around creating systems and technologies that are in harmony with the cognitive abilities of healthcare professionals and patients. This approach aims to deeply understand human cognition—its strengths and limitations—and how these aspects influence the design of interactive systems. A key focus is on user-centered design principles that aim to minimize cognitive load and enhance the clarity of communication. Through the optimization of information presentation and the provision of relevant clinical decision support promptly, cognitive engineering eases more effective decision-making processes.

Moreover, cognitive engineering is instrumental in developing tools that are not only straightforward and easy to use but also supportive of patient needs. This encompasses enhancing the usability of health information systems, such as electronic health records, and aiding in complex cognitive tasks, including the interpretation of medical images. By finding and addressing cognitive biases and strongly emphasizing patient safety, cognitive engineering fosters a synergistic relationship between technology and human cognitive processes, thereby ensuring the delivery of high-quality healthcare.

In the realm of XAI, integrating cognitive and usability engineering principles is paramount. This integration ensures that AI frameworks within clinical settings are designed with a keen awareness of the cognitive demands placed on users. By doing so, these systems not only become more accessible and intuitive for healthcare professionals and patients alike but also promote a safer and more effective healthcare environment. This comprehensive approach underscores the necessity of considering both cognitive and usability aspects in the development and implementation of XAI systems, ensuring they are aligned with the cognitive workflows and needs of their end users [149, 150].

### *8.8. Uncertainty quantification*

This section highlights the significance of assessing and understanding the inherent uncertainties within medical predictions, diagnoses, and treatment outcomes. This aspect becomes increasingly vital with the integration of AI models into healthcare, where the ability to make safe and informed decisions depends on accurately estimating the levels of uncertainty associated with AI-generated insights. [145]. The study underscores the importance of continuous improvement and bias mitigation in AI models to enhance the fairness and equity of healthcare delivery. It advocates for a collaborative approach involving caregivers, data scientists, quality assessment professionals, and ethical boards. This multidisciplinary collaboration ensures that enhancements to AI models are informed by clinical ability and adhere to ethical standards, thereby aligning technological advancements with the core values of healthcare [151].

Furthermore, the establishment of a robust feedback loop with end-users and stakeholders is identified as crucial for finding potential areas for improvement. This engagement helps the refinement of the AI systems, making them more adaptable to the changing dynamics of healthcare needs. Through this iterative feedback and improvement process, AI healthcare applications are poised to become more reliable, exact, and interpretable. Such advancements in AI capabilities are instrumental in improving patient care, decision-making processes, and the overall effectiveness of medical treatments. The emphasis on uncertainty quantification and continuous improvement reflects a commitment to advancing AI technologies in a manner that prioritizes patient safety, treatment efficacy, and the ethical implications of automated decision-making in healthcare [152] [153] [154].

### *8.9.Explainable casual inference techniques*

Causal inference plays a crucial role in the realm of AI, particularly when giving reliable advice in uncertain situations. This process involves determining whether an observed relationship between variables reflects a cause-and-effect dynamic. For AI systems to offer trustworthy recommendations, they must find these relationships and articulate the rationale behind their suggestions, including any associated risks. This capability is achieved through the deployment of causal models, which map out the connections between various actions and their potential outcomes, thereby showing the system's comprehension of the possible consequences of different actions [155].

Moreover, a sophisticated AI system is characterized by its adaptability to changes in a user's intentions and aims. Such a system is adept at recognizing alterations in the user's plans, promptly adjusting its responses based on previously learned patterns or, when necessary, engaging in more deliberate reasoning to learn the most proper course of action. This level of responsiveness and understanding underscores the importance of causal reasoning in AI, enabling it to function effectively across various scenarios. By harnessing the principles of causal inference, AI technologies can advance beyond mere pattern recognition, moving towards a deeper understanding of the intricacies of cause and effect, which is essential for generating insights and advice that are both relevant and reliable [156].

### *8.10. Kaizen and continuous improvement*

Upon a comprehensive evaluation of the XAI process within our system, it becomes clear that continuous improvement stands out as the most crucial element. The philosophy of Kaizen, which focuses on continuous, incremental improvement, offers substantial benefits to healthcare and medicine by enhancing quality, safety, satisfaction, trust, efficiency, and the interpretability of AI systems for end-users. By systematically reducing errors, defects, and variability, the Kaizen method significantly improves the transparency, robustness, and efficiency of the care monitoring process and its outcomes [152, 157].

In the context of XAI in healthcare, the application of Kaizen principles is essential for improving patient outcomes and keeping the trustworthiness of AI-driven systems. This approach entails an iterative cycle of refining models, algorithms, and interpretability techniques informed by real-world feedback and the continuous evolution of medical knowledge. The regular update of models to include new data, research findings, and clinical insights is fundamental, allowing AI systems to remain relevant and effective in changing healthcare demands and landscapes.

Moreover, the accuracy of automated disease diagnosis can be compromised by several factors, including data noise. This highlights the importance of quantifying and communicating the uncertainty inherent in AI models to ensure the reliability of AI diagnoses. By embracing the Kaizen approach within the XAI framework, healthcare organizations can foster a culture of continuous learning and improvement, leading to more reliable, understandable, and user-friendly AI applications that support high-quality patient care and informed clinical decision-making [158].

### *8.11. XAI for wearable health trackers and medical LLMs*

In the healthcare sector, wearable Internet of Things (IoT) devices stand for a significant advancement, serving as patient-worn gadgets that connect directly to the cloud. These devices are instrumental in collecting and transmitting health data in real time, helping with ongoing health monitoring, and enabling rapid responses to medical needs. Wearable technology has proven to be exceptionally beneficial in various aspects of healthcare, including the management of chronic diseases, delivery of therapy, support in rehabilitation, diagnostics, and physical activity tracking [159].

As of 2022, the market for wearable medical devices was estimated at approximately $22.44 billion, projections suggesting it could escalate to around $60.48 billion by 2027. This expected growth, featuring a Compound Annual Growth Rate (CAGR) of 21.9%, is driven by several key factors: the rising incidence of lifestyle-related chronic conditions, an expanding elderly population, increased demand for home healthcare and remote monitoring solutions, and a greater focus on personalized patient care. Notably, 86% of patients have reported experiencing improved health outcomes because of wearable technologies. This surge in demand has led to a growing number of healthcare providers seeking specialized wearable software, which ensures the efficient and secure transmission of Patient-Generated Health Data (PGHD) to cloud-based servers and into Electronic Health Records (EHR) or Electronic Medical Records (EMR) systems. Moreover, such software is equipped with advanced analytics capabilities that further refine healthcare monitoring and decision-making processes [160].

Wearable medical devices, complemented by their associated applications, provide healthcare professionals with a detailed overview of a patient's health status, empowering patients to actively take part in checking their health. This collaborative approach can significantly improve care outcomes. Despite the increasing adoption and clear importance of wearable technology in healthcare, there is still a notable gap in the application of XAI within this domain. Given the substantial growth and reliance on wearable technologies, as highlighted by the statistics, it is imperative to focus on developing and designing a robust, interpretable framework for these devices. Such advancements would enhance the functionality and user experience of wearable medical devices and ensure their decisions are transparent and understandable to patients and healthcare providers, thereby fostering trust and reliability in wearable healthcare technology.

## 9. Conclusion

In conclusion, artificial intelligence (AI) is poised to significantly influence the future of medicine, aiming to maximize patient benefits. Nonetheless, the exclusive reliance on explainability to guide individual patient decisions might not stand for the best strategy. Current explainability methods fall short of guaranteeing the correctness of specific decisions, fostering trust, or justifying the clinical application of AI recommendations. This does not diminish the importance of explainability in ensuring AI safety; it still is an asset for diagnosing issues within models and auditing systems to enhance their performance and uncover potential biases or problems.

For now, explainability methods should serve as instruments for developers and auditors. Unless there are groundbreaking developments in explainable AI, it might be necessary to approach these systems as black boxes, placing trust in them based on their empirically verified performance. The quest for explanations that are understandable to humans in the context of complex machine learning algorithms presents an ongoing challenge. Consequently, healthcare professionals should exercise caution when relying on AI-generated explanations. Similarly, regulators should deliberate carefully before mandating explanations for the clinical implementation of AI technologies.

This study highlights the paramount importance of Trustworthy AI in healthcare, offering a thorough examination of the Explainable AI (XAI) literature landscape. It proves standard definitions and categorizes interpretability across data pre-processing, model selection, and post-processing stages. The suggested roadmap for XAI implementation advocates for setting clear design goals, engaging a diverse range of end-users, and fostering interdisciplinary collaboration. Furthermore, the study tailors its approach to meet the specific needs of various user groups for explanations. It underscores the merging of critical factors such as fairness, privacy, accountability, sustainability, and robustness to create reliable AI pipelines in clinical decision support systems.

Significant strides have been made in enhancing the explainability of ML models through the cooperation of the AI and healthcare communities. However, translating these advancements into practical applications within real-world settings is still essential for building genuine trust in intelligent health systems. Closing the gap between legal requirements and technological progress is crucial for developing risk-aware scenarios and broadening the scope of trustworthiness requirements. While interpretable AI in healthcare aims to foster better understanding and trust, the current methodologies may not suffice to realize fully explainable AI systems. Therefore, interactive systems that offer explanations and ask for feedback play a vital role in convincing end-users and decision-makers of the reliability of AI, easing its adoption without reservations.

**Declaration of competing interest**

The authors declare that they have no known competing commercial interests or personal relationships that could have appeared to influence the work reported in this paper.

## Data availability

Data will be made available on request.

## Acknowledgments

This work did not have any financial support.

*Appendix 1. Abbreviations*

| **IML** | **Interpretability Machine Learning** | **DL** | **Deep Learning** |
|---|---|---|---|
| **XAI** | Explainability Artificial Intelligence | RF | Random Forest |
| **CDSS** | Clinical Decision Support System | DT | Decision Tree |
| **DHTs** | Digital Health Technologies | LR | Logistic Regression |
| **DHIs** | Digital Health Interventions | KNN | K-Nearest Neighbors |
| **IHS** | Intelligent Health System | SVM | Support Vector Machine |
| **HCI** | Human-Computer Interaction | ANN | Artificial Neural Network |
| **WMDs** | Wearable Medical Devices | DNN | Deep Neural Network |
| **MDs** | Medical Devices | CNN | Convolutional Neural Network |
| **WHTs** | Wearable Health Trackers | RNN | Recurrent Neural Network |
| **IoMT** | Internet of Medical Things | NLP | Natural Language Processing |
| **MLLMs** | Medical Large Language Models | VLM | Vision Language Model |
| **mHealth** | Mobile Health | LSTM | Long Short-Term Memory networks |
| **DCTs** | Digital Care Twins | EDA | Exploratory Data Analysis |
| **SDOH** | Social Determinants of Health | PCP | Parallel Coordinate Plots |
| **EMR/EHR** | Electronic Medical/Health Records | LIME | Interpretable Model-agnostic Explanations |
| **WHO** | World Health Organization | SHAP | Shapley Additive Explanations |

*Appendix 2. Criteria for study selection using PRISMA 2020 and PICO.*

| | **Inclusion Criteria** | **Exclusion Criteria** |
|---|---|---|
| **Population** | AI-Based Medical Devices, and Clinical Decision Support System (CDSS) | |
| **Language** | English. | |
| **Study design** | All peer-reviewed and open-access articles in journals and conferences. | Review protocols, and review of review study. |
| **Intervention** | Studies on the XAI and IML in AI-based MDs and CDSS to improve quality and interpretability in intelligent health systems. | Studies that were focusing on:<br>Using classical ML tools |
| **Comparison** | All studies with the comparison of interventions. | |
| **Outcome** | Studies reported results that featured any domain of IML and XAI in AI-based MDs and CDSS as its main finding or as one of the main findings. | |
| **Time limit** | All publication dates were accepted. (2012 - 2023) | Duplicate of the same study. |

*Appendix 3. Results of the structured searching.*

| | Database | Search string | Returned papers |
|---|---|---|---|
| **#1** | PubMed | ("internet of medical things"[tw] OR "IoMT"[tw])<br>("internet of things"[tw] OR "IOT"[tw])<br>("Telehealth"[tw] OR "Telemedicine"[tw] OR "Telecare"[tw] OR "Telemonitoring"[tw])<br>("Remote Health"[tw] OR "Virtual Health"[tw] OR "eHealth"[tw] OR "Digital Health"[tw])<br>("Smart Devices"[tw] OR "Intelligence Devices"[tw] OR "Wearable Medical Device"[tw] OR "Wearable Sensors"[tw])<br>("Digital Twins"[tw])<br>("Medical Large Language Models"[tw] OR “Health Large Language models” [tw] OR “Clinical Large Language Models"[tw])<br>AND (“Explainable Artificial Intelligence"[tw] OR “Interpretable Machine Learning” [tw] OR “Transparent machine learning” [tw] OR “Reliable AI” [tw]) | 658 |
| **#2** | Scopus | ({internet of medical things} OR {IoMT})<br>({internet of things} OR)<br>({Telehealth} OR {Telemedicine} OR {Telecare} OR {Telemonitoring})<br>({Remote Health} OR {Virtual Health} OR {eHealth} OR {Digital Health})<br>({Smart Devices} OR {Intelligence Devices} OR {Wearable Medical Device} OR {Wearable Sensors})<br>({Digital Twins})<br>({Medical Large Language Models} OR {Health Large Language models} OR {Clinical Large Language Models})<br>AND ({Explainable Artificial Intelligence} OR {Interpretable Machine Learning} OR {Transparent machine learning} OR {Reliable AI}) | 1001 |
| **#3** | Web of Science | ("internet of medical things" OR "IoMT")<br>("internet of things" OR "IOT (Internet of Things)")<br>("Telehealth" OR "Telemedicine" OR "Telecare" OR "Telemonitoring")<br>("Remote Health" OR "Virtual Health" OR "eHealth" OR "Digital Health")<br>("Smart Devices" OR "Intelligence Devices" OR "Wearable Medical Devices" OR "Wearable Sensors")<br>("Digital Twins")<br>("Medical Large Language Models" OR “Health Large Language models” OR “Clinical Large Language Models")<br>AND (“Explainable Artificial Intelligence" OR “Interpretable Machine Learning” OR “Transparent machine learning” OR “Reliable AI”) | 582 |
| **#4** | Final | (#1) AND (#2) AND (#3) | 2241 |

*Appendix 4. Summary of important surveys on IML and XAI in healthcare and medicine.*

| | | | | | | | AI-Based Medical Devices | | | | | IML Process | | |
|---|---|---|---|---|---|---|---|---|---|---|---|---|---|---|
| | Ref | Year | Publisher | Type of review | Topic of review | Main field of discipline | Medical LLMs | Telehealth | Smart MWDs | mHealth | Digital Twins | Pre-modeling | Modeling | Post-modeling |
| ***** | **Our Review** | 2023 | | Systematic Review | Robust Interpretability Approach to Enhance Trustworthy AI in Healthcare: A Systematic Review of the Past Decade Toward a Proposed Framework | IML and XAI processes in healthcare, Evaluation, Implementation, and several types of AI-based MDs | * | * | * | * | * | * | * | * |
| **1** | [42] | 2023 | Mayo Clinic | Scoping review | Health Technology Assessment of Performance, Interpretability, and Explainability in AI-based MDs | Guidelines & Healthcare Stakeholders | | * | | | | | * | * |
| **2** | [24] | 2023 | arxiv | Brief Review | A Brief Review of Explainable Artificial Intelligence in Healthcare | XAI methods in healthcare apps | | * | | | | | * | * |
| **3** | [161] | 2023 | Elsevier | Systematic Review | Application of explainable artificial intelligence in medical health | Usability and reliability of XAI | | * | | | | | * | * |
| **4** | [35] | 2023 | Oxford | Narrative review | Precision oncology: a review to assess interpretability in several explainable methods | Comparison of XAI methods in Precision oncology | | * | | | | | * | * |
| **5** | [36] | 2023 | MDPI | Systematic Review | Interpretable ML Techniques in ECG-Based Heart Disease Classification | interpretable ML techniques by focusing on heart disease diagnosis from an ECG signal | | | * | * | | | * | * |
| **6** | [16] | 2023 | arxiv | Systematic Review | Explainable Artificial Intelligence for Drug Discovery and Development--A Comprehensive Survey | XAI methods, their application in drug discovery | | * | | | | | * | * |
| **7** | [162] | 2022 | Science Direct | Systematic Review | Application of explainable AI for healthcare: A systematic review of the last decade (2011–2022) | Comparison of XAI methods in healthcare | | * | | | | | * | * |
| **8** | [43] | 2021 | Elsevier | Narrative review | Opening the Black Box: The Promise and Limitations of Explainable Machine Learning in Cardiology | XAI in Cardiology | | * | | | | | * | * |
| **9** | [163] | 2021 | IEEE | Narrative review | A Survey on Explainable Artificial Intelligence (XAI): Toward Medical XAI | XAI methods in medical | | * | | | | | * | * |
| **10** | [131] | 2020 | Oxford | Scoping review | Explainable artificial intelligence models using real-world electronic health record data | XAI in biomedical and medicine HER | | * | | | | | * | * |

*Appendix 5. Compact summary of existing papers on interpretable intelligent health systems.*

| | | | | | | | Interpretability Process | | |
|---|---|---|---|---|---|---|---|---|---|
| **No** | Ref | Year | Publisher | Topic & Objectives | Application | Methodology & Key Findings | Pre-modeling | Modeling | Post-modeling |
| 1 | [63] | 2023 | Frontiers | XAI-enabled human activity recognition (HAR) method, relies on crucial features extracted from smart home's sensor data | Physical activity for people with chronic impairments | Improved recognition accuracy by Local Interpretable Model Agnostic (LIME) with a Random Forest (RF) classifier. | * | * | * |
| 2 | [18] | 2023 | JMIR | Predicting the Next-Day Perceived and Physiological Stress of Pregnant Women by Using Machine Learning and Explainability | Prenatal stress | Shapley Additive Explanations, Built XAI models & showed risk factors to predict next-day parental stress. | | * | * |
| 3 | [17] | 2023 | JMIR | Explainable Machine Learning Techniques to Predict Amiodarone-Induced Thyroid Dysfunction Risk | Thyroid | Shapley additive explanation (SHAP) a tool for Personalized Risk Prediction and Clinical Decision Support. | | * | * |
| 4 | [123] | 2023 | Springer | Heart Rate Variability-Based Mental Stress Detection: An Explainable Machine Learning Approach | Heart Rate & Mental Stress | SHAP Global Explainability provides more details for physicians by visualizing outcomes. | | * | * |
| 5 | [53] | 2023 | WILEY | Transparent machine learning suggests a key driver in the decision to start insulin therapy in individuals with type 2 diabetes | Diabetes | logic learning machine, supporting evidence-based medicine in real-time data. | * | * | * |
| 6 | [164] | 2023 | INFORM S | Interpretable Policies and the Price of Interpretability in Hypertension Treatment Planning | Hypertension | Markov decision process (MDPs), monotone, clear, efficient, and effective policies for empower decision-making | * | * | * |
| 7 | [165] | 2023 | ECAI | Harnessing the Power of XAI in Prompt-Based Healthcare Decision Support Using ChatGPT | Heart Disease | Contextual prompts, zero-shot and few-shot prompt learning based on LLMs, Important insights on feature significance to support clinical decision-making processes | * | * | * |
| 8 | [126] | 2023 | arXiv | Towards Interpretable Mental Health Analysis with ChatGPT | Mental health | pre-trained language models (PLMs), ChatGPT is an XAI tool. | * | * | * |
| 9 | [97] | 2023 | ICML | A ChatGPT Aided Explainable Framework for Zero-Shot Medical Image Diagnosis | Brain Tumor | Pretrained vision-language models (VLMs), CLIP, effectiveness of VLMs & LLMs for medical applications | * | * | * |
| 10 | [12] | 2023 | SPIE | A user interface to communicate interpretable AI decisions to radiologists | Breast Cancer | Improved the readers' confidence & accuracy in clinical decisions | * | * | * |
| 11 | [68] | 2023 | Duke | Learned Kernels for Interpretable and Efficient PPG Signal Quality Assessment and Artifact Segmentation | Cardiovascular disease | Learned Kernels, Reliable & clear signal assessment in low-power devices | * | * | * |
| 12 | [166] | 2023 | CHIL (Conferen ce on Health, Inference, and Learning) | Missing Values and Imputation in Healthcare Data: Can Interpretable Machine Learning Help? | Infant Death | Explainable Boosting Machines (EBMs) , understand missing data, and risks | * | * | * |
| 13 | [167] | 2023 | Duke | An Interpretable Machine Learning System to Identify EEG Patterns on the Ictal-Interictal-Injury Continuum | Brain injury in ICU | Proto Med-EEG, Increased end-user's trust of ML prediction | * | * | * |
| 14 | [76] | 2023 | Science Direct | Explainable attention ResNet18-based model for asthma detection using stethoscope lung sounds | Asthma | Gradient-weighted class activation mapping (Grad-CAM), obtained high accuracy and explainable results | * | * | * |
| 15 | [52] | 2023 | Science Direct | Adazd-Net: Automated adaptive and explainable Alzheimer's disease detection system using EEG signals | Alzheimer | LIME, SHAP, and MS, Adaptive Flexible Analytic Wavelet Transform (AFAWT), precise and explainable AZD detection | * | * | * |
| 16 | [77] | 2023 | MDPI | Explainable Risk Prediction of Post-Stroke Adverse Mental Outcomes Using Machine Learning Techniques in a Population of 1780 Patients | Post-stroke adverse mental outcome (PSAMO) | Shapley additive explanations (SHAP), early-stage intervention after a stroke. | * | * | * |
| 17 | [71] | 2023 | MDPI | An Explainable Deep Learning Model to Prediction Dental Caries Using Panoramic Radiograph Images | Dental Caries | ResNet-50, better performance compares to EfficientNet-B0 and DenseNet-121 | * | * | * |
| 18 | [62] | 2022 | arXiv | Automatic Diagnosis of Myocarditis Disease in Cardiac MRI Modality Using Deep Transformers and Explainable AI | Myocarditis Disease | Turbulence Neural Transformer (TNT), Explainable-based Grad Cam method | * | | * |
| 19 | [55] | 2022 | Science Direct | Explainable multi-module semantic guided attention-based network for medical image segmentation | multi-organ CT images, Brain tumor MRI images | multi-module semantic guided attention-based network (MSGA-Net), high performance of all metrics | * | * | * |
| 20 | [168] | 2022 | Duke | Interpretable Deep Learning Models for Better Clinician-AI Communication in Clinical Mammography | Mammography | Margin predictions & shape predictions, More explainable perdition models | * | * | * |
| 21 | [118] | 2022 | IEEE | XAIoT framework to track physiological health with the smartwatch. | Physiological health | Ease hospital crowding, offer clear medical explanations for each patient, and faster diagnosis by tracking individually. | | | * |
| 22 | [119] | 2022 | Springer | Utilizing fog computing and explainable deep learning techniques for gestational diabetes prediction | Gestational Diabetes | Shapley additive explanation (SHAP) to provide local & global explanation and early prediction with cost cost-effective solution. | * | * | * |
| 23 | [169] | 2022 | Elsevier | Neural Additive Models for Explainable Heart Attack Prediction | Heart Attack | Neural Additive Models, evaluate prediction efficiency | | | * |
| 24 | [170] | 2022 | IEEE | Enhancing interpretability & usability of local factors in decision-making | Child Welfare Screening | SIBYL, an interpretable & interactive visualization tool for predictive algorithms | | * | * |
| 25 | [96] | 2022 | PubMed | Consistency of Feature Importance Algorithms for Interpretable EEG Abnormality Detection | EEG Abnormality | Created CNN using multiple EEG frequencies as input & 4 methods for feature importance: LRP (Layer wise Relevance Propagation) (Layer wise Relevance Propagation) (Layer wise Relevance Propagation), Deep LIFT, IG, & Guided Grad CAM. | | | * |
| 26 | [171] | 2022 | PubMed | Enhancing XAI by examining subgroups of features within a ML model | Alzheimer | Harvard-Oxford (Subcortical) Atlas, evaluating expert definition categories in decision making. | | | * |
| 27 | [172] | 2022 | Nature | Interpretability and fairness evaluation of deep learning models on the MIMIC-IV dataset | Mortality prediction | IMVLSTM69. Simplified Interpretative model for learning network parameters & feature importance at the same time | * | * | * |
| 28 | [173] | 2022 | PLOS | Application of early detection mortality or unplanned readmission in a retrospective cohort study | Unplanned Death | Shapley variable selection tool (Shapley VIC), Enhancing Interpretability of Prediction Models for Decision Makers | | * | * |
| 29 | [174] | 2022 | Frontiers | Prediction of Online Psychological Help-Seeking Behavior During the COVID-19 Pandemic: An Interpretable ML | Mental Health | Shapley additive explanation (SHAP), Enable quick, early, and easily understandable detection | | * | * |
| 30 | [175] | 2022 | MDPI | Logic Learning Machine-Based Explainable Rules Accurately Stratify the Genetic Risk of Primary Biliary Cholangitis | Primary Biliary Cholangitis | logic learning machine (LLM), an efficient individual-level predictive tool with interpretable rule extraction. | * | * | * |
| 31 | [176] | 2022 | MDPI | Measuring the Usability and Quality of Explanations of a ML Web-Based Tool for Oral Tongue Cancer Prognostication | Oral Tongue Cancer | System Usability Scale (SUS) and system Causality Scale (SCS), Provide deliverable usability & explainable outcome | | * | * |
| 32 | [177] | 2022 | Frontiers | Machine Learning Approaches for Hospital Acquired Pressure Injuries: A Retrospective Study of Electronic Medical Records | Hospital-Acquired Pressure Injuries | Shapley additive explanation (SHAP), real-time predictive methods to prevent unnecessary harms | * | * | * |
| 33 | [178] | 2022 | IEEE | A New XAI-based Evaluation of Generative Adversarial Networks for IMU Data Augmentation | Physical fatigue monitoring | Logic Learning Machine aids in finding the fake dataset | * | * | * |
| 34 | [179] | 2022 | PMC | Opening the black box: interpretable machine learning for predictor finding of metabolic syndrome | Metabolic syndrome | Post-hoc model-agnostic interpretation methods found the most reliable predictors. | | | * |
| 35 | [180] | 2022 | PMC | Finding the presence and severity of dementia by applying interpretable ML techniques on structured clinical records | Dementia | Decision trees, correct prediction of the existence and severity of dementia disease | * | * | * |
| 36 | [181] | 2022 | PMC | Explainable machine learning methods and respiratory oscillometers for the diagnosis of respiratory abnormalities in sarcoidosis | Respiratory diseases | Genetic Programming & Grammatical Evolution aided clinician decision-making and enhanced pulmonary function service productivity. | * | * | * |
| 37 | [182] | 2022 | PMC | Application of machine learning techniques for predicting survival in ovarian cancer | ovarian cancer | SHAP method + DT + RF, more reliable & transparency | * | * | * |

| | | | | | | | | | | |
|---|---|---|---|---|---|---|---|---|---|---|
| 38 | [183] | 2021 | Nature | Passive detection of COVID-19 with wearable sensors & XAI tools | COVID-19 | Explainable gradient boosting prediction model based on decision trees, Highlighted applicability in Settings without Self-Reported symptoms from any device. | * | * | * |
| 39 | [184] | 2021 | Nature | A case-based interpretable deep learning model for the classification of mass lesions in digital mammography | Breast Cancer | interpretable AI algorithm for breast lesions, more exact and explainable even with the small size of the image dataset | * | * | * |
| 40 | [185] | 2021 | IEEE | An automated feature selection and classification pipeline to improve the explainability of clinical prediction models | Kidney, cardiovascular, diabetes | Ensemble Trees ML models, Pipeline to find the best feature selection and ML methods | * | * | * |
| 41 | [19] | 2021 | Nature | Human activity recognition using wearable sensors, discriminant analysis, and long short-term memory-based neural structured learning | Elderly people | Local Interpretable Model-Agnostic Explanations (LIME), Detect people's behavior in different settings | * | | * |
| 42 | [186] | 2021 | MDPI | Early Detection of Septic Shock Onset Using Interpretable Machine Learners | Septic Shock | 8 ML models, diagnosis with real-time clinical & administrative data within the first 6 hours of admission | * | * | * |
| 43 | [187] | 2021 | ESC | "Explainable AI in Cardiology": A Tool to Provide Personalized Predictions on Cardiac Health States Among Older Adults Engaged in Physical Activity | Cardiology | Role of XAI tools like Shapley Addictive Explanations (SHAP) to measure the feature importance in practical clinical applications. | | * | * |
| 44 | [187] | 2021 | Springer | Predicting the Probability of 6-Month Unfavorable Outcome in Patients with Ischemic Stroke | Ischemic Stroke | Global & local interpretability techniques, death reduction of stroke by aiding caregivers in the decision-making process | | * | * |
| 45 | [20] | 2021 | Sensors | Find the connection between collected heart features (HRV & BPM) from medical devices & COVID-19 by interpretable AI | Covid-19 | Infection prediction 48 hours (about 2 days) prior to any symptoms by Local Interpretable Model-Agnostic Explanations (LIME). | * | * | * |
| 46 | [121] | 2021 | Sensors | Augment data from IoMT Sensors with GAN, & use XAI to compare with real data | Chronic Obstructive Pulmonary Disease | GAN, LLM, DT, Comparable generated data with real date | | * | * |
| 47 | [101] | 2021 | Springer | An Explainable Artificial Intelligence Framework for the Deterioration Risk Prediction of Hepatitis Patients | Hepatitis & liver disease | Shapley additive explanations (SHAP), Local Interpretable Model-agnostic Explanations (LIME), and partial Dependence Plots (PDP), improve the transparency & decision-making of complex models | * | * | * |
| 48 | [188] | 2021 | BMC | SMILE: systems metabolomics using interpretable learning and evolution | Alzheimer's disease | linear genetic programming (LGP), clear and easy-to-understand predictive models | | | * |
| 49 | [80] | 2021 | BMC | Patient similarity analytics for explainable clinical risk prediction | type-2 diabetes mellitus, hypertension & dyslipidemia | K-nearest neighbor, Practical method to develop more explainable ML | | * | * |
| 50 | [90] | 2021 | PubMed | A New Approach for Interpretability and Reliability in Clinical Risk Prediction: Acute Coronary Syndrome Scenario | Acute coronary syndrome (ACS) | logistic regression (LR), artificial neural network (ANN), and clinical risk score model (namely the Global Registry of Acute Coronary Events - GRACE) | | * | * |
| 51 | [2] | 2020 | ACM | Diabetes detection by using data from smartwatch sensors | Diabetes | SHAP (Shapley Additive Explanations) to attribute features & explain decisions clearly on smartwatches. | * | * | * |
| 52 | [95] | 2020 | PMC | Interpretable Predictions of Clinical Outcomes with an Attention-based Recurrent Neural Network | Predict mortality in the MIMIC-III database | Recurrent Neural Network, Higher accuracy and more interpretable by visualization | | | * |
| 53 | [189] | 2020 | IOS Press | Machine Learning Explainability to predict 10-year overall survival of breast cancer | Breast Cancer | SHapley Additive explanations (SHAP), Local Interpretable Model-agnostic Explanations (LIME), better acceptance of AI tools | | | * |
| 54 | [190] | 2020 | MDPI | Explainable Machine Learning Framework for Image Classification Problems | Glioma Cancer | Compare White Box vs Black Box models, with transparent feature extraction method, & explainable prediction framework | * | * | * |
| 55 | [91] | 2020 | PLOS | Opening the black box of artificial intelligence for clinical decision support | Stroke | Tree boosting & multilayer perceptron (MLPs), modern ML methods can provide more explainability. | * | | * |
| 56 | [191] | 2020 | PMC | Ada-WHIPS: explaining AdaBoost classification with applications in the health sciences | Computer Aided Diagnostics (CAD) | Adaptive-Weighted High Importance Path Snippets (Ada-WHIPS), proved a better explanation | * | * | * |
| 57 | [102] | 2019 | PMC | On the interpretability of a machine learning-based model for predicting hypertension | Hypertension | 5 global interpretability techniques (Feature Importance, Partial Dependence Plot, Individual Conditional Expectation, | * | * | * |
| | | | | | | Feature Interaction, Global Surrogate Models) & 2 local interpretability techniques (Local Surrogate Models, Shapley) | | | |
| 58 | [192] | 2018 | Elsevier | Enhancing interpretability of automatically extracted machine learning features: application to RBM-Random Forest system on brain lesion segmentation | Brain tumor & Ischemic stroke | RBM-Random Forest, Easy to understand, highlights the pattern | | * | * |
| 59 | [193] | 2016 | Nature | MediBoost: a Patient Stratification Tool for Interpretable Decision Making in the Era of Precision Medicine | Precision Medicine | MediBoost, a single & easy to understand tree with high accuracy which like ensemble methods | | * | * |
| 60 | [194] | 2016 | PubMed | Interpretable Deep Models for ICU Outcome Prediction | Acute lung injury (ALI) | Gradient boosting trees Explainable results for clinicians | | * | * |
| 61 | [81] | 2015 | Elucid | INTERPRETABLE CLASSIFIERS USING RULES AND BAYESIAN ANALYSIS: BUILDING A BETTER STROKE PREDICTION MODEL | STROKE | Bayesian Rule Lists, more exact & easier to understand | | * | * |
| 62 | [195] | 2014 | PubMed | Interpretable Associations over DataCubes: application to hospital managerial decision making | Breast Cancer | COGARE method, simpler rules, and enhanced explainability improved overfitting | | * | * |
| 63 | [144] | 2014 | PLOS | Ant Colony Optimization Algorithm for Interpretable Bayesian Classifiers Combination | Heart Disease & Cardiography | Ant Colony Optimization, as well as Bagging and boosting models, improved communication with patients | * | | * |
| 64 | [143] | 2012 | PLOS | A Mathematical Model for Interpretable Clinical Decision Support | Gynecology | Interval coded scoring (ICS) system, Enhanced patient-clinician communication | | * | * |